\documentclass[journal]{IEEEtran}
\usepackage{cite}
\usepackage[T1]{fontenc}
\usepackage{amsmath,amssymb}
\usepackage{caption}
\usepackage{array}
\usepackage{graphicx}
\usepackage{float}
\usepackage{tabularx}
\usepackage[hidelinks]
{hyperref}
\usepackage{booktabs}
\usepackage{multirow}
\begin{document}

\title{State-Dependent Memristor Dynamics for Device-Informed Spike-Timing-Dependent Plasticity in Spiking Neural Networks}

\author{
Samiur Rahman Khan$^{1}$ and Saifur Rahman Sabuj$^{2}$\\[4pt]
\small $^{1}$University of Cincinnati, Cincinnati, OH, USA\\
\small $^{2}$BRAC University, Dhaka, Bangladesh\\
\small Email: Khan4sn@mail.uc.edu
}

\maketitle

\maketitle

\begin{abstract}
This work investigates a device-informed learning framework in which local spike-timing-dependent plasticity (STDP) commands determine programming-pulse polarity and a programming coordinate that drives a literature-derived threshold-memristor state update. The first fully connected layer is trained using local STDP coupled to state-dependent conductance updates, whereas a supervised backpropagation-trained readout layer performs class assignment. HSPICE simulations characterize thresholded, polarity-dependent, and state-dependent programming behavior of the adopted compact model and analyze differential conductance mapping and static signed multiply--accumulate operation in a reduced crossbar classifier. Network-level experiments on MNIST and Fashion-MNIST compare the hybrid update with direct clipped STDP and evaluate controlled STDP update-amplitude asymmetry, persistent faults during training, selected device-parameter variation, stochastic write noise, and pixel-level input corruption. The Hybrid model achieved 95.15\% accuracy on MNIST and 85.92\% on Fashion-MNIST. In a separate seed-matched comparison, Hybrid reached 95.02\% and 84.79\%, compared with 92.33\% and 83.25\% for direct clipped STDP, respectively. Potentiation-dominant asymmetry caused a rapid loss of input-dependent feature selectivity on MNIST, while accuracy under endpoint faults present during training remained substantially higher than under the same faults injected after training. The network-level mapping is a validation-selected, reduced-order device-informed state-update abstraction that retains the threshold, polarity, current, and state dependence of the adopted compact model; it does not constitute a pulse-accurate physical programming implementation. The results show that, under the evaluated compact-model mapping and training protocol, the device-informed state-update transformation alters
local-STDP learning behavior, and they characterize the resulting sensitivity to the controlled nonidealities considered here.
\end{abstract}


\section{Introduction}
\label{sec:introduction}

Spiking neural networks (SNNs) use temporally sparse spike events and local
neural dynamics for neuromorphic sensing, inference, and online adaptation
\cite{mead1990neuromorphic,maass1997networks,indiveri2015memory,furber2016large}.
Nonvolatile resistive-memory devices can encode synaptic conductances and
support analog current summation in crossbar arrays, motivating
device-informed neural computation
\cite{strukov2008missing,jo2010nanoscale,prezioso2015training,
burr2017neuromorphic,ielmini2018inmemory,sebastian2020memory}.
Their conductance can be programmed by electrical pulses, enabling
synapse-like learning rules such as spike-timing-dependent plasticity (STDP)
\cite{jo2010nanoscale,serrano2013stdp}.

Physical resistive-memory programming, however, is generally thresholded,
nonlinear, state dependent, polarity asymmetric, and subject to device and
cycle variability~\cite{burr2015experimental,yu2018neuroinspired}.
Many SNN studies instead employ ideal weight updates or
abstract nonideality models that do not explicitly capture threshold
activation, finite conductance range, or state-dependent potentiation and
depression
\cite{serrano2013stdp,querlioz2013immunity,agarwal2016resistive,
sung2018review,sayyaparaju2020deviceaware}.
Consequently, STDP behavior under direct or bounded weight updates may differ
when the same learning commands are transformed through thresholded,
state-dependent device dynamics.

Memristive hardware has demonstrated STDP learning and broader neural-network
computing~\cite{pedretti2017memristive,prezioso2018spike,wang2018fully,
wang2025realtime}. Device-aware circuit studies have examined STDP switching
\cite{serrano2013stdp,sayyaparaju2020deviceaware,zhang2026hardware}, while
network-level work has examined nonlinear updates, variation, homeostasis, and
faults~\cite{kim2021snn,querlioz2013immunity,vu2024nonidealities,
prajapati2025efficient}. These studies establish
the importance of incorporating device behavior into neuromorphic learning
and motivate examining how explicit state-dependent switching dynamics alter
STDP at the network level.

This work couples local STDP to the literature-derived device dynamics of
Zhang \emph{et al.}~\cite{zhang2017memristive}. Batch-aggregated STDP commands
are converted into a dimensionless programming coordinate and polarity that
drive a thresholded, state- and current-dependent compact-model update.
Kim \emph{et al.}~\cite{kim2021snn} studied nonlinear LTP/LTD updates in a
memristive SNN; here, local STDP is coupled to the adopted threshold-, state-,
and current-dependent compact-model dynamics. Unlike prior circuit-level
device-aware STDP studies
\cite{serrano2013stdp,sayyaparaju2020deviceaware}, network-level variation
and homeostasis studies \cite{querlioz2013immunity}, and supervised
memristive-network fault studies \cite{prajapati2025efficient}, the present
work evaluates how this state-dependent transformation affects learning
relative to direct clipped STDP through a response-factor ablation, a
pulse-amplitude sweep across the switching thresholds, and
systematic evaluation of update asymmetry, persistent
faults, device-parameter variation, write noise, and input corruption.

Complementary HSPICE simulations characterize the adopted compact model and
evaluate static signed multiply--accumulate (MAC) behavior in a reduced
differential crossbar under
specified mismatch and read-voltage perturbations; this circuit analysis
provides a consistency check of differential conductance mapping and static
current summation, distinct from the full \(784\times1024\) network
implementation.

The main contributions are as follows:
\begin{itemize}

\item A hybrid device-informed SNN in which local fc1 STDP commands determine
programming polarity and a dimensionless programming coordinate for
thresholded, state-dependent memristor updates, with a supervised fc2 readout
for class assignment.

\item A controlled comparison of the device-informed update with direct
clipped STDP, accompanied by a response-factor ablation, a pulse-amplitude
sweep across the switching thresholds, and a normalization ablation, together
with evaluation of \(A_+/A_-\)
update-amplitude asymmetry, training-time and post-training endpoint faults,
device-parameter variation, stochastic write noise, and input corruption.

\item HSPICE characterization of the adopted compact device model and reduced
differential-crossbar signed-MAC operation under device mismatch,
read-voltage perturbation, and interconnect resistance.

\end{itemize}

\section{Device Model and Characterization}
\label{sec:device-model-characterization}

\subsection{Threshold-Memristor Model}
\label{subsec:threshold-memristor-model}

The voltage-controlled threshold-memristor compact model and
Ag/AgInSbTe/Ta parameter set reported by Zhang et
al.~\cite{zhang2017memristive} provide the device basis for the circuit- and
network-level simulations. The device resistance is expressed in terms of the
internal state variable \(w(t)\) as

\begin{equation}
R(t)=
R_{\mathrm{ON}}\frac{w(t)}{D}
+
R_{\mathrm{OFF}}
\left(
1-\frac{w(t)}{D}
\right),
\label{eq:memristance}
\end{equation}
where \(D\) is the state-variable range, \(w_{\mathrm{norm}}=w/D\in[0,1]\)
is the normalized state, and \(R_{\mathrm{ON}}\) and
\(R_{\mathrm{OFF}}\) denote the limiting low- and high-resistance states,
respectively. State evolution follows the polarity-dependent,
threshold-gated relation

\begin{equation}
\frac{dw(t)}{dt}
=
\begin{cases}
\displaystyle
\mu_v
\frac{R_{\mathrm{ON}}}{D}
\frac{i_{\mathrm{off}}}{i(t)-i_0}
f\!\left(w(t)\right),
&
v(t)>V_{\mathrm{TH}}^+>0,
\\[8pt]
0,
&
V_{\mathrm{TH}}^- \leq v(t) \leq V_{\mathrm{TH}}^+,
\\[6pt]
\displaystyle
\mu_v
\frac{R_{\mathrm{ON}}}{D}
\frac{i(t)}{i_{\mathrm{on}}}
f\!\left(w(t)\right),
&
v(t)<V_{\mathrm{TH}}^-<0,
\end{cases}
\label{eq:state_dynamics}
\end{equation}
with the Joglekar boundary window function~\cite{joglekar2009elusive}

\begin{equation}
f\!\left(w(t)\right)
=
1-
\left(
2\frac{w(t)}{D}-1
\right)^{2p},
\label{eq:window_function}
\end{equation}
where \(p\) controls suppression of state evolution near the resistance
limits. State evolution is zero within
\(V_{\mathrm{TH}}^- \leq v(t) \leq V_{\mathrm{TH}}^+\).
The adopted parameters are listed in Table~\ref{tab:aist_parameters}.

\begin{table}[h]
\centering
\caption{Parameters of the adopted threshold-memristor model
\cite{zhang2017memristive}.}
\label{tab:aist_parameters}
\begin{tabular}{lc}
\hline
Parameter & Value \\
\hline
\(R_{\mathrm{ON}}\) & \(10~\Omega\) \\
\(R_{\mathrm{OFF}}\) & \(420~\Omega\) \\
\(V_{\mathrm{TH}}^-\) & \(-0.19~\mathrm{V}\) \\
\(V_{\mathrm{TH}}^+\) & \(0.37~\mathrm{V}\) \\
\(D\) & \(3~\mathrm{nm}\) \\
\(\mu_v\) & \(1.6\times10^{-12}~\mathrm{m^2\,s^{-1}\,\Omega^{-1}}\) \\
\(i_{\mathrm{on}}\) & \(1~\mathrm{A}\) \\
\(i_{\mathrm{off}}\) & \(10^{-5}~\mathrm{A}\) \\
\(i_0\) & \(10^{-3}~\mathrm{A}\) \\
\(p\) & 5 \\
\hline
\end{tabular}
\end{table}

Under the selected polarity convention, positive programming drives the
device toward \(R_{\mathrm{ON}}\) and higher conductance, whereas negative
programming drives it toward \(R_{\mathrm{OFF}}\) and lower conductance.
The network-level mapping uses \(\pm0.6~\mathrm{V}\), exceeding the
corresponding thresholds \(V_{\mathrm{TH}}^+=0.37~\mathrm{V}\) and
\(V_{\mathrm{TH}}^-=-0.19~\mathrm{V}\). At this amplitude, the minimum
device current, \(0.6~\mathrm{V}/R_{\mathrm{OFF}}\approx1.43~\mathrm{mA}\),
exceeds \(i_0\), so the positive-branch denominator in
Eq.~\eqref{eq:state_dynamics} remains positive. In general, the positive
branch requires \(v>i_0R(w)\), which can fail only for
\(V_{\mathrm{TH}}^{+}<v<i_0R(w)\leq0.42~\mathrm{V}\).

\subsection{Compact-Model Characterization}
\label{subsec:device-characterization}

\begin{figure*}[!t]
\centering
\includegraphics[width=\textwidth]{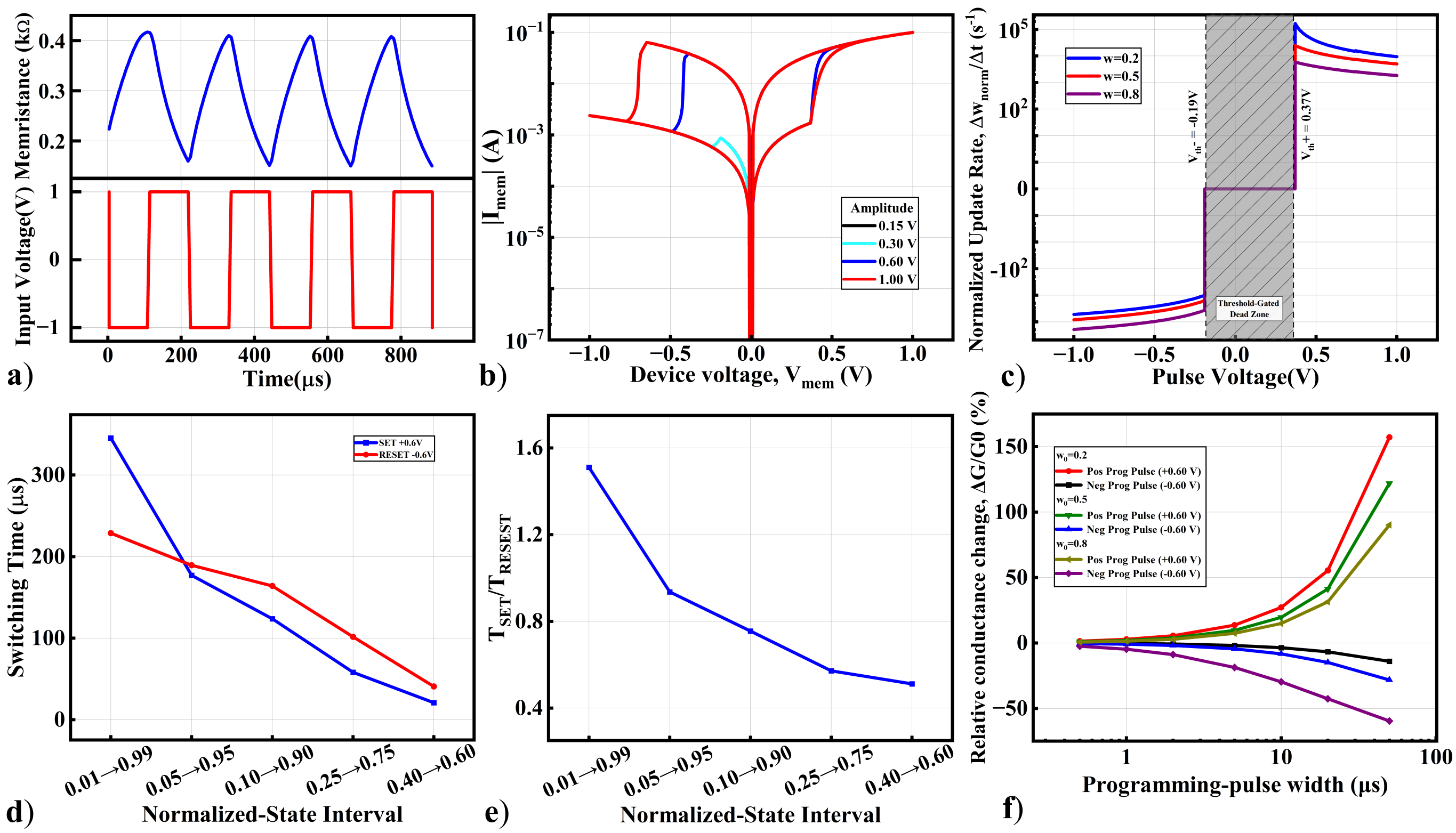}
\caption{HSPICE characterization of the adopted threshold-memristor model:
(a) transient memristance response under alternating
\(\pm1.0~\mathrm{V}\), \(110~\mu\mathrm{s}\) programming pulses;
(b) current--voltage response under 100-Hz sinusoidal excitation with peak
voltages from \(0.15\) to \(1.00~\mathrm{V}\);
(c) average normalized state-update rate
\(\Delta w_{\mathrm{norm}}/\Delta t\) over a 100-ns pulse versus applied
voltage for \(w_{\mathrm{norm},0}=0.2\), \(0.5\), and \(0.8\), on a
symmetric logarithmic vertical axis with the threshold dead zone shaded;
(d) SET (\(+0.6~\mathrm{V}\)) and RESET (\(-0.6~\mathrm{V}\)) transition
times over normalized-state intervals;
(e) corresponding \(T_{\mathrm{SET}}/T_{\mathrm{RESET}}\) ratios; and
(f) relative conductance change versus programming-pulse width for
\(w_{\mathrm{norm},0}=0.2\), \(0.5\), and \(0.8\) under
\(\pm0.60~\mathrm{V}\) pulses.}
\label{fig:Device_Model_Characterization}
\end{figure*}

Transient HSPICE simulations of the Verilog-A compact model characterize the
device dynamics underlying the network-level mapping; the implementation
evaluates Eq.~\eqref{eq:state_dynamics} directly, with zero velocity only
for \(\lvert i-i_0\rvert<10^{-15}~\mathrm{A}\), and clips the state to
\([0,D]\).
Figure~\ref{fig:Device_Model_Characterization} summarizes the programming,
threshold, switching-kinetic, and state-dependent responses.

\subsubsection{Transient Programming Response}

Figure~\ref{fig:Device_Model_Characterization}(a) shows the memristance
response to alternating \(\pm1.0~\mathrm{V}\), \(110~\mu\mathrm{s}\)
programming pulses. Positive and negative pulses drive memristance toward
\(R_{\mathrm{ON}}\) and \(R_{\mathrm{OFF}}\), respectively. The incomplete
single-pulse transitions and reduced response near the resistance limits
reflect finite-duration programming and the boundary behavior of
Eq.~\eqref{eq:window_function}.

\subsubsection{Sinusoidal Excitation Response}

Figure~\ref{fig:Device_Model_Characterization}(b) shows the current--voltage
response under 100-Hz sinusoidal excitation. At \(0.15~\mathrm{V}\), the
excitation remains within the threshold interval and produces negligible
state evolution. At \(0.30~\mathrm{V}\), only \(V_{\mathrm{TH}}^-\) is
crossed, producing negative-polarity-selective switching, whereas
\(0.60~\mathrm{V}\) and \(1.00~\mathrm{V}\) exceed both thresholds and
produce bipolar hysteretic responses, with an effective SET onset from the
high-resistance boundary at \(i_0R_{\mathrm{OFF}}\approx0.42~\mathrm{V}\).

\subsubsection{Threshold Dead Zone}

Figure~\ref{fig:Device_Model_Characterization}(c) shows the average
normalized state-update rate \(\Delta w_{\rm norm}/\Delta t\), obtained from
the state change over a 100-ns pulse, versus applied voltage for
\(w_{\rm norm,0}=0.2\), \(0.5\), and \(0.8\). State evolution is negligible
within the threshold interval
\(-0.19~\mathrm{V}\leq v\leq0.37~\mathrm{V}\), whereas outside this range
the update rate depends on voltage polarity and initial state; the sharp
rise just above \(V_{\mathrm{TH}}^{+}\) reflects the
\(i_{\mathrm{off}}/(i-i_0)\) term as the device current approaches \(i_0\).
The
\(+0.10~\mathrm{V}\) crossbar read voltage lies within this dead zone,
enabling modeled conductance sensing without intentionally activating state
evolution.

\subsubsection{Range-Dependent Switching Kinetics}

Figures~\ref{fig:Device_Model_Characterization}(d) and
\ref{fig:Device_Model_Characterization}(e) compare SET and RESET transition
times under matched \(+0.6~\mathrm{V}\) and \(-0.6~\mathrm{V}\) programming.
Transition times were evaluated over matched normalized-state intervals,
from lower to upper state for SET and upper to lower state for RESET.
The ratio \(T_{\mathrm{SET}}/T_{\mathrm{RESET}}\) varies from \(1.51\) over
\(0.01\)--\(0.99\) to \(0.51\) over \(0.40\)--\(0.60\), demonstrating that
SET/RESET asymmetry depends on the evaluated state range.

\subsubsection{State-Dependent Conductance Modulation}

Figure~\ref{fig:Device_Model_Characterization}(f) shows the relative
conductance change,
\(\Delta G/G_0=(G_{\mathrm{final}}-G_0)/G_0\), produced by single
\(\pm0.60~\mathrm{V}\) pulses for \(w_{\mathrm{norm},0}=0.2\), \(0.5\),
and \(0.8\). Update magnitude depends on initial state, pulse polarity, and
duration. At \(w_{\mathrm{norm},0}=0.2\), a positive
\(50~\mu\mathrm{s}\) pulse produces approximately \(+157\%\) conductance
change compared with approximately \(-14\%\) for negative programming.
At \(w_{\mathrm{norm},0}=0.8\), the corresponding changes are approximately
\(+90\%\) and \(-60\%\), respectively.

Since the SET/RESET relation varies with device state and operating range,
the LTP/LTD coefficient asymmetry examined in
Section~\ref{subsec:stdp-amplitude-asymmetry} is treated as a controlled
algorithmic perturbation rather than a single device-intrinsic asymmetry
constant.

\subsection{Reduced Crossbar Consistency Analysis}
\label{subsec:crossbar-mac-wire}

A reduced \(64\times10\) differential crossbar was simulated in HSPICE to
evaluate conductance-based signed multiply--accumulate (MAC) operation and
read-path sensitivity using the threshold-memristor compact model. The array
implements a bias-free linear classifier for a synthetic ten-class benchmark
of hand-drawn \(8\times8\) binary digit glyphs, providing 64 input features
and 10 output classes while remaining tractable for transient circuit
simulation. Each signed weight is represented by a differential pair of
non-negative conductances, \(G_{ij}^{+}\) and \(G_{ij}^{-}\), as commonly
used for signed-weight encoding~\cite{prezioso2015training}, such that
\begin{equation}
\Delta G_{ij}=G_{ij}^{+}-G_{ij}^{-}.
\label{eq:crossbar_diff_conductance}
\end{equation}

This differential representation is used only for the signed-MAC circuit
analysis; the hybrid SNN fc1 layer instead uses the unipolar
conductance-derived mapping in
Eq.~\eqref{eq:network-conductance-weight}. The \(64\times10\) logical array
requires 20 physical output columns and 1280 device instances. Active and
inactive input rows were driven at \(+0.10~\mathrm{V}\) and \(0~\mathrm{V}\),
respectively, with the positive and negative output columns held at virtual
ground by ideal current-sense sources. The \(+0.10~\mathrm{V}\) read voltage
lies within the threshold dead zone identified in
Section~\ref{subsec:device-characterization}, enabling conductance sensing
without intentionally activating state evolution. The differential class
score and predicted class are

\begin{align}
s_j&=I_{\mathrm{sense},j}^{+}-I_{\mathrm{sense},j}^{-}
\propto
\sum_{i=1}^{64}x_i
\left(G_{ij}^{+}-G_{ij}^{-}\right),
\label{eq:crossbar_score} \\
\hat{y}&=\operatorname*{argmax}_{j}(s_j).
\label{eq:crossbar_prediction}
\end{align}

The reduced crossbar experiment used a synthetic ten-class digit-glyph
benchmark rather than MNIST. Each class was represented by a fixed
\(8\times8\) binary template. The training set contained 3,000 noisy glyphs
(300 per class) with pixel-inversion probability
\(p_{\rm noise}^{\rm train}=0.10\), and the test set contained 150 glyphs
(15 per class) with \(p_{\rm noise}^{\rm test}=0.18\).

A bias-free linear softmax classifier was trained from zero-initialized
weights by full-batch gradient descent for 5,000 iterations with a learning
rate of 0.35 and an \(\ell_2\) penalty coefficient of \(10^{-4}\). The
absence of a bias term allows its numerical class scores to be represented
directly by Eq.~\eqref{eq:crossbar_score}. The base random seed was 20260810,
with derived seeds \(+1\), \(+2\), and \(+3\) for device mismatch,
state checking, and read-voltage perturbation, respectively. The trained
weights were normalized and mapped to symmetric differential conductance
pairs,
\begin{equation}
G_{ij}^{+}
=
G_{\rm mid}
+
sW_{ij},
\qquad
G_{ij}^{-}
=
G_{\rm mid}
-
sW_{ij},
\label{eq:differential-weight-mapping}
\end{equation}
where
\(G_{\rm mid}=(G_{\max}+G_{\min})/2\),
\(s=0.90(G_{\max}-G_{\min})/2\),
\(W_{ij}\leftarrow W_{ij}/\max_{i,j}|W_{ij}|\),
\(G_{\min}=1/R_{\mathrm{OFF}}\), and
\(G_{\max}=1/R_{\mathrm{ON}}\). 

For a binary template pixel $x_i\in\{0,1\}$, the noisy input pixel
$\tilde{x}_i$ was generated as
\begin{equation}
\tilde{x}_i
=
x_i\oplus m_i,
\qquad
m_i\sim\operatorname{Bernoulli}(p_{\rm noise}),
\label{eq:glyph-pixel-noise}
\end{equation}
where $\oplus$ denotes XOR and $p_{\rm noise}$ equals
$p_{\rm noise}^{\rm train}=0.10$ for training samples and
$p_{\rm noise}^{\rm test}=0.18$ for test samples.

The normalized weights were mapped once before circuit evaluation, and the
resulting conductances were held fixed during all reads. The 150-image test
used fixed, device-specific 5\% Gaussian variation in \(R_{\mathrm{ON}}\)
and \(R_{\mathrm{OFF}}\), with sampled parameters fixed across all reads;
samples violating \(R_{\mathrm{ON}}<0.5R_{\mathrm{OFF}}\) were clipped to
that bound. Active rows additionally received independent zero-mean Gaussian
read-voltage perturbations with an unclipped standard deviation of
\(20~\mathrm{mV}\), clipped to the threshold dead zone.


This circuit-level mismatch experiment is distinct from the 5--30\%
network-level device-parameter-variation study in
Section~\ref{subsec:fault-variation-robustness}, which also perturbs
\(V_{\mathrm{TH}}^{+}\) and \(V_{\mathrm{TH}}^{-}\) during training and
evaluation.

\begin{figure}[H]
\vspace{-6pt}
\centering
\includegraphics[width=\columnwidth,height=4.9cm]{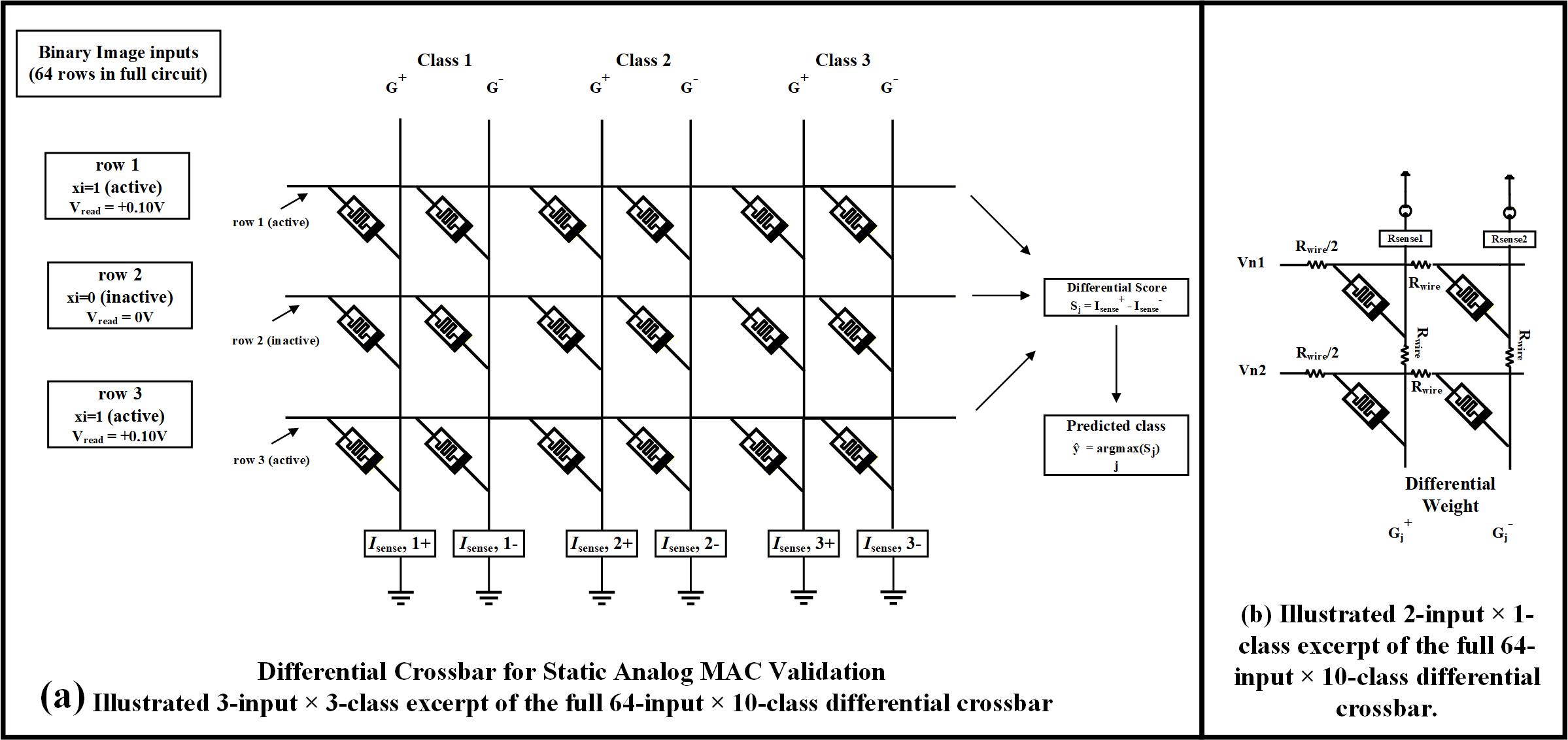}
\caption{Reduced differential-crossbar HSPICE analysis.
(a) Differential conductance mapping and static signed-MAC configuration,
shown as a three-input, three-class excerpt of the \(64\times10\) logical
classifier, with each signed weight represented by the differential pair
\(G_{ij}^{+}\) and \(G_{ij}^{-}\).
(b) Interconnect model used for the resistance sweep, shown as a two-input,
one-class excerpt: input rows are driven through
\(R_{\mathrm{wire}}/2\), row and column segments include
\(R_{\mathrm{wire}}\), and each physical output column (\(G_j^{+}\) and
\(G_j^{-}\)) connects to ideal current sensing through
\(R_{\mathrm{sense}}\), with \(R_{\mathrm{sense}}=2~\Omega\) (fixed) or
\(2R_{\mathrm{wire}}\) (proportional).}
\label{fig:crossbar-mac-wire}
\vspace{-6pt}
\end{figure}

Under combined device mismatch and read-voltage perturbation, the HSPICE
array correctly classified 145 of 150 noisy glyphs (96.67\%), the same
overall accuracy as the nominal numerical classifier, with predicted classes
matching for 146 of 150 images (97.33\%) and a Pearson score correlation of
0.9777. When the same sampled device parameters and row-voltage
perturbations were applied to the numerical reference, the
\(150\times10=1500\) image--class scores showed a Pearson correlation of
1.0000 with HSPICE and 100\% winner agreement. This matched-condition
comparison shows consistent differential conductance mapping and static
current summation between the HSPICE and numerical implementations. No state
change was detected in the 16 monitored devices across the 150 reads (2400
device-read observations).

Line resistance causes position-dependent IR drop in crossbar
arrays~\cite{liang2010crosspoint,chen2013comprehensive}.
Interconnect sensitivity was evaluated for the same trained classifier under
nominal device and read-voltage conditions on a 30-image test set (3 per
class) using the interconnect model of
Fig.~\ref{fig:crossbar-mac-wire}(b) over
\(R_{\mathrm{wire}}=\{10^{-3},\allowbreak 0.02,\allowbreak 0.05,\allowbreak
0.10,\allowbreak 0.20,\allowbreak 0.30,\allowbreak 0.50,\allowbreak
0.70,\allowbreak 1.0,\allowbreak 1.5,\allowbreak 2.0\}~\Omega/\mathrm{segment}\),
with either fixed
\(R_{\mathrm{sense}}=2~\Omega\) or proportional
\(R_{\mathrm{sense}}=2R_{\mathrm{wire}}\) sensing. The ideal-sensing HSPICE
reference matched the numerical accuracy of 93.33\%. With proportional
sensing, accuracy was 93.33\%, 90.00\%, 80.00\%, and 60.00\% at
\(R_{\mathrm{wire}}=10^{-3}\), 0.02, 0.05, and
\(0.10~\Omega/\mathrm{segment}\), respectively. With fixed \(2~\Omega\)
sensing, the corresponding accuracies were 90.00\%, 93.33\%, 86.67\%, and
73.33\%. Both configurations reached 26.67\% at
\(1.0~\Omega/\mathrm{segment}\); on this test set, one image corresponds to
3.33 percentage points. For fixed sensing, the mean near-to-far
voltage drop along an active row increased from 0.28~mV at
\(10^{-3}~\Omega/\mathrm{segment}\) to 22.7~mV at 0.10 and 68.0~mV at
\(1.0~\Omega/\mathrm{segment}\), while the ideal--nonideal Pearson score
correlation decreased from 0.955 to 0.689 and 0.350, respectively. No
state change was detected in the monitored devices at any sweep point,
supporting interconnect IR drop rather than read disturbance as the source
of the observed degradation.

\subsection{Paired-Pulse and Algorithmic Timing Responses}
\label{subsec:paired-pulse-stdp}

Paired pre- and post-synaptic programming waveforms were applied to the
adopted threshold-memristor model in HSPICE over the relative timing offset
\(\Delta t=t_{\mathrm{post}}-t_{\mathrm{pre}}\). Starting from \(w_{\mathrm{norm},0}=0.5\), the physical timing response was
calculated as
\(\Delta w_{\mathrm{phys}}=100(w_{\mathrm{norm,final}}-
w_{\mathrm{norm},0})\).
For comparison, the STDP update response was evaluated over the same timing
offsets; its accumulated update determines
programming polarity and the dimensionless programming coordinate \(q\) in
the reduced-order state mapping.

\begin{figure}[t]
\centering
\includegraphics[width=\columnwidth]{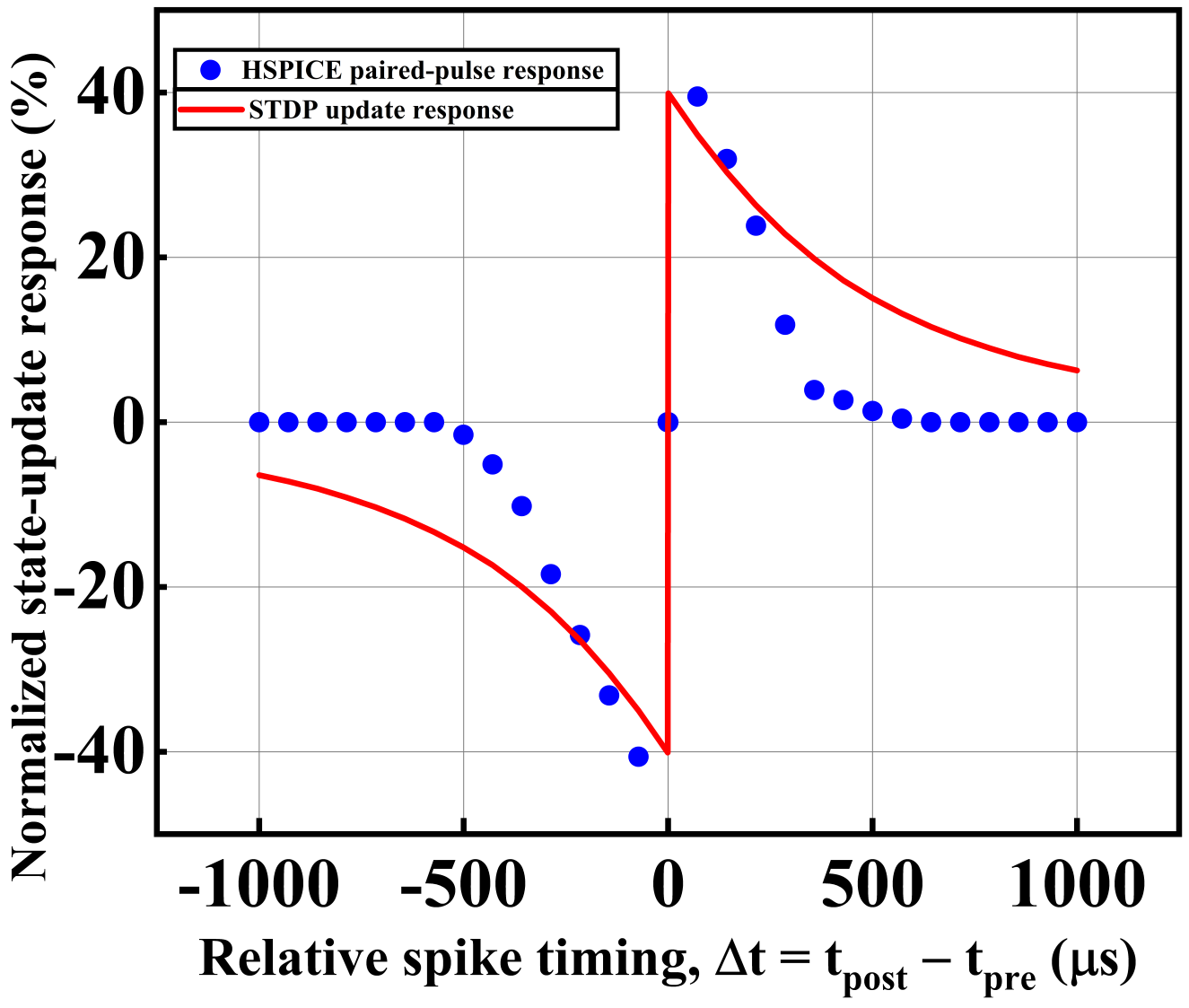}
\caption{Qualitative timing comparison between the HSPICE paired-pulse
response and the trace-based STDP timing response, shown versus relative
timing
\(\Delta t=t_{\rm post}-t_{\rm pre}\) in \(\mu\mathrm{s}\).}
\label{fig:paired-pulse-stdp}
\end{figure}

Figure~\ref{fig:paired-pulse-stdp} shows potentiation for positive
\(\Delta t\) and depression for negative \(\Delta t\), with decreasing
update magnitude as \(|\Delta t|\) increases. The HSPICE response,
\(\Delta w_{\mathrm{phys}}\), results from transient integration of paired
voltage waveforms, whereas the algorithmic response,
\(\Delta w_{\mathrm{algo}}\), represents the trace-based STDP timing
response. Their common timing grid therefore provides a qualitative
comparison of update polarity and timing trend rather than amplitude or
time constant. The HSPICE response decays more steeply toward zero, and
the paired-pulse response is not used in the network-level simulations.

\section{Network-Level Learning Framework}
\label{sec:network-framework}

Network-level simulations were implemented in PyTorch using a reduced-order
realization of the adopted threshold-memristor state dynamics. The hybrid SNN
uses device-informed local STDP for fc1 hidden-feature learning and a
supervised fc2 readout for class assignment.

\subsection{Hybrid Network Architecture and Data Protocol}
\label{subsec:hybrid-network}

The architecture consists of an input-to-hidden layer fc1 and a linear
hidden-to-output classifier fc2. We refer to this network as Hybrid
because it combines device-informed STDP learning in fc1 with a
backpropagation-trained fc2 readout. The fc1 layer is trained by local
STDP coupled to the threshold-memristor state-update model, whereas fc2
is trained at each mini-batch by backpropagation with the Adam optimizer
and cross-entropy loss, while fc1 is updated exclusively through the
local STDP-driven state updates described below.

MNIST and Fashion-MNIST were used for
evaluation~\cite{lecun1998gradient,xiao2017fashion}, with \(28\times28\)
grayscale images providing \(n_{\mathrm{in}}=784\) input features. MNIST training and test sets were class balanced by undersampling each class
to the count of the smallest class (digit 5), giving 5,421 training images
per class (54,210 in total) and 892 test images per class (8,920 in total).
A separate class-balanced subset was drawn for each random seed, and
conditions evaluated with the same seed used the same subsets. A separate
MNIST experiment used the native 60,000-image training and 10,000-image test
sets for the final configuration. Fashion-MNIST used its native balanced 60,000-image training
set and 10,000-image test set. Pixel intensities were clipped to \([0,1]\)
and rate encoded over \(T\) timesteps, using Bernoulli spikes with
probability equal to the normalized pixel intensity during both training
and inference.

At each timestep, rate-coded inputs drive the hidden layer through the normalized conductance-derived matrix \(\mathbf{W}_{\mathrm{fc1}}\). Hidden neurons follow current-based leaky-integrate-and-fire dynamics with
subtractive reset~\cite{neftci2019surrogate}: the membrane potential decays with a leak factor of 0.9, integrates the weighted input, and is reduced by one threshold unit after crossing the unit firing threshold. Winner-take-all competition retains the \(k\) highest-potential threshold-crossing neurons for each sample and timestep; if fewer than \(k\) neurons cross threshold, all threshold-crossing neurons are retained. Only retained spikes are
emitted and trigger the subtractive reset; nonretained neurons carry their
membrane potential forward. The spike-count vector accumulated
over \(T\) timesteps forms the hidden representation supplied to fc2 for
class prediction.

For the \(1024\)-hidden-neuron configurations, \(T=25\) and \(k=24\) were
used for MNIST, and \(T=15\) and \(k=32\) for Fashion-MNIST. Each input
therefore requires \(Tn_{\mathrm{in}}n_{\mathrm{hidden}}\) fc1 MAC
operations, corresponding to \(20{,}070{,}400\) for MNIST and
\(12{,}042{,}240\) for Fashion-MNIST.

Local STDP is applied only to fc1. Pre- and post-synaptic traces follow
exponentially decaying recurrences \cite{serrano2013stdp,indiveri2015memory},

\begin{equation}
\begin{aligned}
x_{\mathrm{pre}}[t] &=
\left(1-\frac{1}{\tau_{\mathrm{pre}}}\right)
x_{\mathrm{pre}}[t-1]+s_{\mathrm{pre}}[t],\\
y_{\mathrm{post}}[t] &=
\left(1-\frac{1}{\tau_{\mathrm{post}}}\right)
y_{\mathrm{post}}[t-1]+s_{\mathrm{post}}[t],
\end{aligned}
\label{eq:stdp-traces}
\end{equation}

where \(\tau_{\mathrm{pre}}=\tau_{\mathrm{post}}\) in all experiments.
Dataset-specific trace constants and the balanced coefficient
\(A_{+}=A_{-}=A_0\) are listed in
Table~\ref{tab:implementation-parameters}. The trace constants are defined
in simulation timesteps, with \(1-1/\tau\) as the corresponding discrete
decay factor. The signed local update at timestep \(t\) is

\begin{equation}
\Delta\mathbf{W}_{\mathrm{raw}}[t]
=
A_{+}\mathbf{x}_{\mathrm{pre}}^{\mathsf{T}}[t]
\mathbf{s}_{\mathrm{post}}[t]
-
A_{-}\mathbf{s}_{\mathrm{pre}}^{\mathsf{T}}[t]
\mathbf{y}_{\mathrm{post}}[t].
\label{eq:stdp-update}
\end{equation}

Here, the pre- and post-synaptic traces and spikes are represented as row
vectors, so the products in Eq.~\eqref{eq:stdp-update} form
\(n_{\mathrm{in}}\times n_{\mathrm{hidden}}\) outer products;
\(\mathbf{s}_{\mathrm{post}}[t]\) denotes the WTA-filtered hidden-layer spike;
WTA is applied before STDP, so only spikes of the \(k\) selected
threshold-crossing neurons enter \(\mathbf{s}_{\mathrm{post}}\) and the
post-synaptic traces. Because both traces include the current-timestep
spike, coincident pre- and post-synaptic spikes contribute
\((A_{+}-A_{-})\,\mathbf{s}_{\mathrm{pre}}^{\mathsf{T}}[t]\mathbf{s}_{\mathrm{post}}[t]\),
which vanishes for balanced coefficients but not under the asymmetric
settings of Section~\ref{subsec:stdp-amplitude-asymmetry}.

The trace-based contributions are accumulated over the \(T\)-timestep
window for each sample and averaged over the mini-batch to obtain the
batch-aggregated command \(\overline{\Delta W}_{\mathrm{raw}}^{b}\).
For each synapse \((i,j)\), the sign of
\(\overline{\Delta W}_{{\rm raw},ij}^{b}\) selects the programming polarity,
and its magnitude determines the programming coordinate in
Eq.~\eqref{eq:effective-duration}. The compact-model state equation then
produces one equivalent state-dependent update per synapse and mini-batch,
with the programming coordinate

\begin{equation}
q_{ij}^{b}
=
S_{\rm dur}
\left|
\overline{\Delta W}_{{\rm raw},ij}^{b}
\right|,
\label{eq:effective-duration}
\end{equation}
where \(q_{ij}^{b}\) is a dimensionless programming coordinate and
\(S_{\mathrm{dur}}\) is a validation-selected, dataset-specific scale.
The coordinate \(q_{ij}^{b}\) specifies update magnitude, while the sign of
\(\overline{\Delta W}_{{\rm raw},ij}^{b}\) selects programming polarity; it
is a reduced-order numerical coordinate, not a physical pulse duration,
pulse energy, or directly measured programming quantity. Before the
state update is evaluated, \(q_{ij}^{b}\) is limited to a maximum value
of 2000.

The reduced-order mapping evaluates the compact-model switching velocity at the pre-update normalized device state, normalizes that velocity in implementation units, and applies the dimensionless programming coordinate determined by the aggregated local command. The mapping preserves the threshold gating, polarity dependence, current dependence, and boundary-window dependence of the adopted compact model while providing a tractable device-state representation of learning at network scale. Let $w_{\mathrm{norm},ij}^{b}\in[0,1]$ be the normalized device state immediately before batch update $b$, and let $v_{ij}=V_{\mathrm{pulse}}$ be the programming voltage. The pre-update
resistance and current are

\begin{equation}
R_{ij}^{b}
=
R_{\mathrm{ON}}w_{\mathrm{norm},ij}^{b}
+
R_{\mathrm{OFF}}\left(1-w_{\mathrm{norm},ij}^{b}\right),
\qquad
i_{ij}^{b}
=
\frac{v_{ij}}{R_{ij}^{b}}.
\label{eq:preupdate-current}
\end{equation}

The compact-model switching velocity is evaluated at the pre-update
normalized state:

\begin{equation}
v_{{\rm sw},ij}^{b}
=
\begin{cases}
\displaystyle
\frac{\mu_v R_{\rm ON}}{D}
\frac{i_{\rm off}}{i_{ij}^{b}-i_0}
f\!\left(w_{{\rm norm},ij}^{b}\right),
&
v_{ij}>V_{\mathrm{TH}}^{+},
\\[8pt]
0,
&
V_{\mathrm{TH}}^{-}\leq v_{ij}\leq V_{\mathrm{TH}}^{+},
\\[8pt]
\displaystyle
\frac{\mu_v R_{\rm ON}}{D}
\frac{i_{ij}^{b}}{i_{\rm on}}
f\!\left(w_{{\rm norm},ij}^{b}\right),
&
v_{ij}<V_{\mathrm{TH}}^{-},
\end{cases}
\label{eq:compact-response-factor}
\end{equation}
where $[v_{\rm sw}]=\mathrm{m\,s^{-1}}$ and the normalized boundary
window is
\begin{equation}
f\!\left(w_{{\rm norm},ij}^{b}\right)
=
1-
\left(
2w_{{\rm norm},ij}^{b}-1
\right)^{2p}.
\label{eq:normalized-window}
\end{equation}

The network-level response coefficient is obtained by normalizing the
compact-model switching velocity with the reference velocity
$v_{\rm ref}$:
\begin{equation}
r_{ij}^{b}
=
\frac{v_{{\rm sw},ij}^{b}}{v_{\rm ref}},
\qquad
[v_{\rm ref}]=\mathrm{m\,s^{-1}}.
\label{eq:normalized-response-coefficient}
\end{equation}
For the network-level implementation,
\(v_{\mathrm{ref}}=1\,\mathrm{m/s}\) is a fixed nondimensionalization
reference, and the factor \(1/D\) in
\(dw_{\mathrm{norm}}/dt=v_{\mathrm{sw}}/D\) is absorbed into
\(S_{\mathrm{dur}}\).
Accordingly, \(S_{\mathrm{dur}}\) is validation selected rather than
calibrated to the pulse-width-dependent conductance response in
Fig.~\ref{fig:Device_Model_Characterization}(f). The mapping thus retains
the compact model's state-, current-, polarity-, and threshold-dependent
switching behavior without imposing a quantitative pulse-width calibration.

The explicit batchwise state update is
\begin{equation}
w_{{\rm norm},ij}^{b+1}
=
\operatorname{clip}\!\left(
w_{{\rm norm},ij}^{b}
+
r_{ij}^{b}q_{ij}^{b},
\,0,\,1
\right).
\label{eq:single-step-update}
\end{equation}

Equation~\eqref{eq:single-step-update} defines the deterministic mapping;
in nominal Hybrid training, each increment \(r_{ij}^{b}q_{ij}^{b}\) is
additionally perturbed before clipping by the update-proportional write
noise of Section~\ref{subsec:write-noise-robustness} with
\(\sigma_{\mathrm{write}}=0.018\).

The single-step approximation evaluates resistance, current, and response
coefficient once at the pre-update state. Its numerical accuracy was tested
against a finer-grained evaluation of the same reduced-order mapping, in
which \(q\) was subdivided and the resistance, current, boundary window, and
response coefficient were re-evaluated after each increment. The 462-point
grid comprised 21 initial states over
\(w_{\rm norm}\in[0.02,0.98]\), both programming polarities at
\(V_{\rm pulse}=\pm0.6~\mathrm{V}\), and 11 values over
\(q\in[0.01,20]\). Across all 462 conditions, the maximum
normalized-state error was \(1.27\times10^{-5}\), and the maximum relative
conductance error was \(2.79\times10^{-4}\). This comparison covers
\(q\leq20\); in full nominal training runs (one seed per dataset), the
largest realized value was \(q=2.02\) on MNIST and \(q=0.93\) on
Fashion-MNIST, below this upper limit.

\begin{figure}[!t]
\centering
\includegraphics[width=\columnwidth]{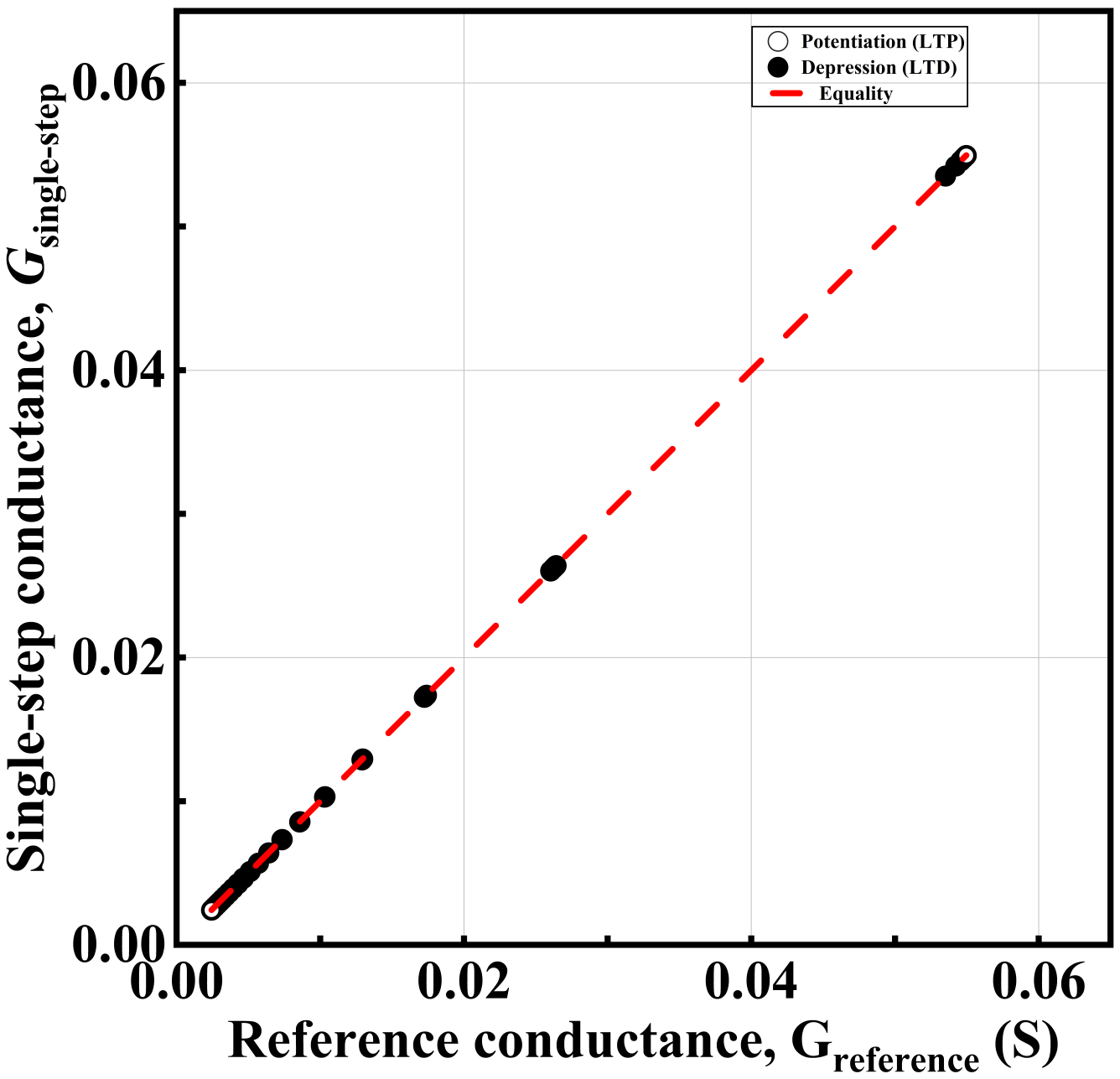}
\caption{Numerical consistency between the explicit single-step update in
Eq.~\eqref{eq:single-step-update} and the finer-grained evaluation of the
same reduced-order mapping across 462 conditions. The dashed line denotes
equality.}
\label{fig:single-step-validation}
\end{figure}

Figure~\ref{fig:single-step-validation} confirms close agreement between the
single-step and finer-grained evaluations of the reduced-order mapping; this
comparison evaluates numerical approximation accuracy, not transient HSPICE
agreement. The updated state determines resistance and conductance through
Eq.~\eqref{eq:memristance}, with \(w_{\mathrm{norm}}=w/D\). The normalized
conductance used as the fc1 forward weight is

\begin{equation}
\begin{aligned}
W_{\mathrm{fc1}}
&= \frac{G-G_{\min}}{G_{\max}-G_{\min}},
&\qquad G&=\frac{1}{R},\\
G_{\min}
&= \frac{1}{R_{\mathrm{OFF}}},
& G_{\max}&=\frac{1}{R_{\mathrm{ON}}}.
\end{aligned}
\label{eq:network-conductance-weight}
\end{equation}

The same normalized device state therefore determines both the
state-dependent programming response and the unipolar
conductance-derived fc1 forward weight
\(W_{\mathrm{fc1}}\in[0,1]\). Because this representation is nonnegative,
hidden-layer selectivity is shaped by leaky-integrate-and-fire dynamics and
winner-take-all competition rather than signed synaptic cancellation.

At the end of each training epoch, motivated by weight-normalization
approaches in STDP networks~\cite{diehl2015unsupervised}, the stored
normalized states feeding each hidden neuron are rescaled column-wise as
\begin{equation}
w_{\mathrm{norm},ij}
\leftarrow
\operatorname{clip}\!\left[
w_{\mathrm{norm},ij}
\frac{0.5n_{\mathrm{in}}}
{\max\!\left(\sum_{i'=1}^{n_{\mathrm{in}}}w_{\mathrm{norm},i'j},10^{-8}\right)},
\,0,\,1
\right].
\label{eq:column-normalization}
\end{equation}
Each hidden-neuron input column is thereby rescaled toward a target
pre-clipping state sum of \(0.5n_{\rm in}\). This epoch-wise column
rescaling is an algorithmic homeostatic normalization applied once at the
end of each training epoch; it modifies the stored states from which the
conductance-derived forward weights are computed and is not part of the
physical memristor state-update model.

For the final \(1024\)-hidden-neuron, batch-128 configurations retrained on
the complete training sets, fc1 contains \(784\times1024=802{,}816\)
synapses. Batchwise update activity was quantified by counting synapses with
nonzero batch-aggregated local commands,
\(\overline{\Delta W}_{{\rm raw},ij}^{b}\neq0\), during training
(Table~\ref{tab:update-activity}). On average, \(52.81\%\) of fc1
synapses received a nonzero command per mini-batch on MNIST, compared with
\(88.50\%\) on Fashion-MNIST.

\begin{table}[!t]
\centering
\caption{Realized fc1 batch-aggregated update activity for the final
1024-hidden-neuron, batch-128 configurations retrained on the complete
training sets. A synapse is counted as active when
\(\overline{\Delta W}_{{\rm raw},ij}^{b}\neq0\). Values are mean
\(\pm\) standard deviation over mini-batches pooled across five independent
seeds.}
\label{tab:update-activity}
\footnotesize
\renewcommand{\arraystretch}{1.10}
\setlength{\tabcolsep}{3pt}
\resizebox{\columnwidth}{!}{%
\begin{tabular}{lcccc}
\hline
Dataset & Synapses & Updated/batch & Active fraction & Batch range \\
\hline
MNIST &
802,816 &
\(423{,}987\pm10{,}291\) &
\(52.81\pm1.28\%\) &
367,678--457,198 \\
Fashion-MNIST &
802,816 &
\(710{,}470\pm20{,}156\) &
\(88.50\pm2.51\%\) &
577,363--755,556 \\
\hline
\end{tabular}%
}
\end{table}

Table~\ref{tab:implementation-parameters} summarizes the network,
programming, and optimization parameters for the final
1024-hidden-neuron, batch-128 hybrid SNN configurations. Dataset-dependent
hyperparameters, including the trace constant, balanced STDP coefficient,
programming-coordinate scale \(S_{\rm dur}\), fc2 Adam learning rate,
and related discrete settings, were selected using Optuna's Tree-structured
Parzen Estimator sampler over 120 trials per dataset, maximizing accuracy on
a validation partition containing 15\% of the training data. The final
configurations were retrained on the complete training sets (54,210 balanced
MNIST and 60,000 Fashion-MNIST images) and evaluated
across \(n=5\) independent seeds using the final-epoch models after the full
100-epoch training schedule.

\begin{table}[t]
\caption{Network and learning parameters for the final hybrid SNN
configurations. Dataset-dependent parameters were selected using validation
data.}
\label{tab:implementation-parameters}
\centering
\renewcommand{\arraystretch}{1.10}
\begin{tabularx}{\columnwidth}{p{0.50\columnwidth}XX}
\toprule
Parameter & MNIST & Fashion-MNIST \\
\midrule
Hidden neurons & 1024 & 1024 \\
Batch size & 128 & 128 \\
Training epochs & 100 & 100 \\
Simulation timesteps, $T$ & 25 & 15 \\
STDP trace constant, $\tau_{\mathrm{pre}}=\tau_{\mathrm{post}}$
(timesteps) & 1.2474 & 1.9955 \\
Balanced STDP coefficient, $A_{+}=A_{-}=A_0$ &
$4.860\times10^{-4}$ & $1.505\times10^{-4}$ \\
Winner-take-all count, $k$ & 24 & 32 \\
LIF leak factor & 0.9 & 0.9 \\
LIF firing threshold & 1.0 & 1.0 \\
Programming voltage, $V_{\mathrm{pulse}}$ &
$\pm0.6\,\mathrm{V}$ &
$\pm0.6\,\mathrm{V}$ \\
Boundary-window exponent, $p$ & 5 & 5 \\
Initial normalized state &
$\mathcal{U}(0.3,0.7)$ &
$\mathcal{U}(0.3,0.7)$ \\
Programming-coordinate scale, \(S_{\mathrm{dur}}\) &
\(4.38\times10^{4}\) &
\(2.16\times10^{4}\) \\
fc2 Adam learning rate & 0.000528 & 0.000665 \\
Write-noise coefficient, \(\sigma_{\mathrm{write}}\) & 0.018 & 0.018 \\
\bottomrule
\end{tabularx}
\end{table}

\section{Network-Level Results}
\label{sec:network-results}

Results from separate experiments correspond to independently trained
populations under the protocol stated for each experiment; clean-reference
accuracies are therefore compared within each experiment.

\subsection{Hybrid Classification Performance}
\label{subsec:hybrid-performance}

The final 1024-hidden-neuron, batch-128 Hybrid configuration achieved
\(95.15\pm0.17\%\) on the 8,920-image class-balanced MNIST test subsets and
\(85.92\pm0.23\%\) on the native 10,000-image Fashion-MNIST test set over
five seeds. In a separate experiment, the same configuration trained on the
native 60,000-image MNIST training set achieved \(95.93\pm0.12\%\) on the
native 10,000-image test set.

Initial normalized device states were evaluated using
\(w_{\mathrm{norm},0}\sim\mathcal{U}(0,1)\),
\(\mathcal{U}(0.3,0.7)\), and \(\mathcal{U}(0.45,0.55)\), with the range
selected on validation accuracy. In this separate five-seed initialization
study, the selected \(\mathcal{U}(0.3,0.7)\) range yielded
\(95.07\pm0.17\%\) on MNIST and \(85.69\pm0.37\%\) on Fashion-MNIST,
compared with \(92.75\pm0.18\%\) and \(84.66\pm0.25\%\) for
\(\mathcal{U}(0,1)\) and \(92.36\pm0.25\%\) and \(84.40\pm0.31\%\) for
\(\mathcal{U}(0.45,0.55)\).

A separate five-seed sensitivity sweep scaled \(S_{\rm dur}\) by
\(0.25\times\), \(0.5\times\), \(2\times\), and \(4\times\) relative to
its selected value. On the class-balanced MNIST protocol
(54,210 training and 8,920 test images), the corresponding accuracies were
\(95.39\pm0.17\%\), \(95.23\pm0.24\%\), \(94.20\pm0.27\%\), and
\(88.13\pm0.42\%\). On Fashion-MNIST, the sensitivity sweep used
51,000 training images and the native 10,000-image test set, yielding
\(85.59\pm0.29\%\), \(85.64\pm0.37\%\), \(85.52\pm0.14\%\), and
\(84.79\pm0.26\%\), respectively. Accuracy was similar at
\(0.25\times\) and \(0.5\times\) on MNIST and from \(0.25\times\) to
\(2\times\) on Fashion-MNIST, while larger scales reduced accuracy,
most strongly at \(4\times\) on MNIST.

\begin{figure}[!t]
\centering
\includegraphics[width=\columnwidth]{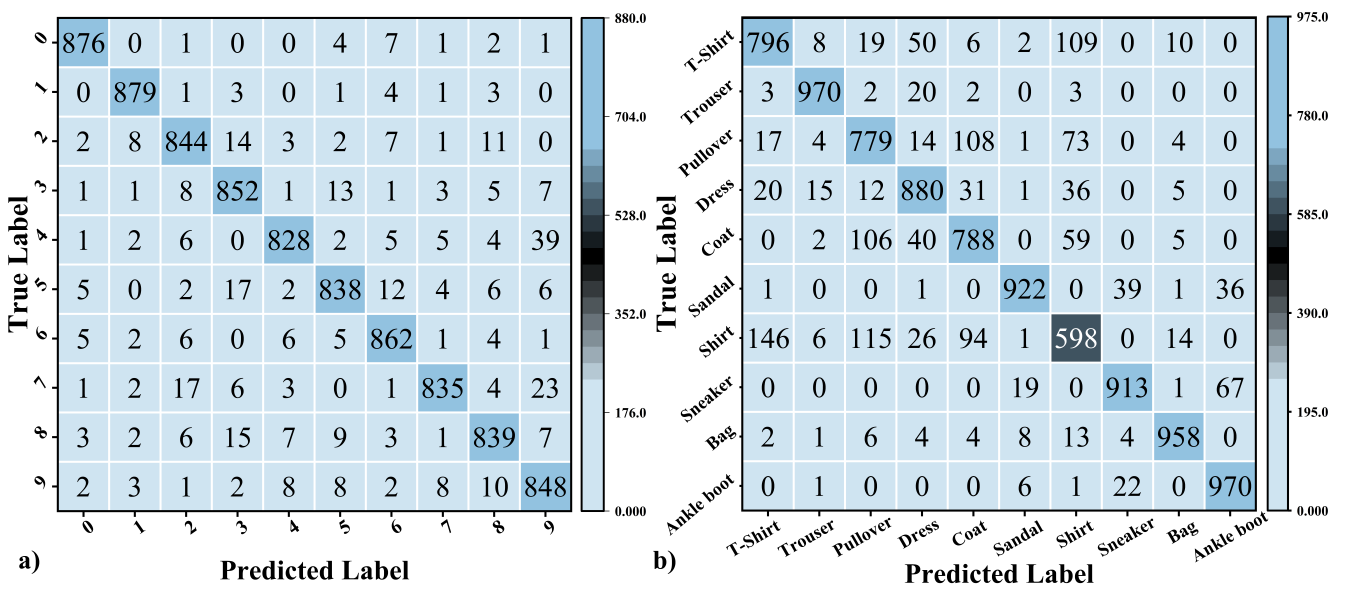}
\caption{Representative confusion matrices for the 1024-hidden-neuron,
batch-128 hybrid SNN: (a) MNIST evaluated on the 8,920-image test set
containing 892 images per class and (b) Fashion-MNIST evaluated on its
native 10,000-image test set containing 1,000 images per class.}
\label{fig:snn_confusion_together}
\end{figure}

Figure~\ref{fig:snn_confusion_together} shows representative
confusion matrices for the 1024-hidden-neuron, batch-128 configuration.

\subsection{Learning-Rule Comparison}
\label{subsec:learning-rule-comparison}

The hybrid SNN (Hybrid) was compared with three references. Ideal STDP
applies the same batch-aggregated local STDP command formula directly to the normalized fc1 weight, without the device-state transformation. In
Ideal BP, fc1 and fc2 are trained end-to-end by surrogate-gradient
backpropagation under the same rate coding, LIF dynamics, WTA, and
100-epoch schedule. For Ideal BP and Memristor BP, the forward spike
function remains the hard threshold used by Hybrid, while its derivative
during backpropagation is approximated by the fast-sigmoid
surrogate~\cite{neftci2019surrogate,zenke2018superspike},
\begin{equation}
\frac{\partial s}{\partial u}
\approx
\frac{1}
{\left(1+10\left|u-V_{\rm th}\right|\right)^2},
\label{eq:surrogate-gradient}
\end{equation}
where \(u\) is the pre-threshold membrane potential and \(V_{\rm th}=1\).
WTA remains a hard top-\(k\) selection in the forward pass. During
backpropagation, the surrogate gradient is retained for the selected
winners and masked for suppressed neurons. Ideal BP uses signed,
unconstrained fc1 weights initialized as
\(\mathrm{fc1}\sim\mathcal{U}(-1/\sqrt{n_{\mathrm{in}}},
1/\sqrt{n_{\mathrm{in}}})\) and trained with Adam without weight decay
or column normalization.

In Memristor BP, fc1 is realized by the same threshold-memristor
array as Hybrid, with initial normalized states drawn from
\(\mathcal{U}(0.3,0.7)\) and forward weights obtained through
Eq.~\eqref{eq:network-conductance-weight}. The surrogate gradient is
computed with respect to the normalized device state through this mapping.
At each mini-batch, Adam generates a state increment whose sign sets the
programming polarity and whose magnitude, multiplied by a programming
scale, sets the programming coordinate in
Eq.~\eqref{eq:effective-duration}. The resulting update is determined by
the same compact-model state equation, threshold gating, write noise, and
state bounds as in Hybrid. Memristor BP was evaluated with and without the
epoch-wise column normalization in Eq.~\eqref{eq:column-normalization}.

For Ideal STDP, the fc1 update is
\begin{equation}
W_{{\rm fc1},ij}^{b+1}
=
\operatorname{clip}\!\left(
W_{{\rm fc1},ij}^{b}
+
\eta\,\overline{\Delta W}_{{\rm raw},ij}^{b},
\,0,\,1
\right),
\label{eq:ideal-stdp-update}
\end{equation}
where \(\overline{\Delta W}_{{\rm raw},ij}^{b}\) is the same
batch-aggregated local command used by Hybrid and \(\eta\) is the
direct-update scale. A separate ten-point log-spaced validation search over
\(10^{-4}\) to \(3.0\) selected \(\eta=0.30\) for MNIST and
\(\eta=0.001\) for Fashion-MNIST. Ideal STDP weights were initialized from
\(\mathcal{U}(0.3,0.7)\) directly in weight space, column normalization
acted on these weights, and no write noise was applied. Ideal STDP uses the normalized state
directly as the forward weight, while Hybrid applies the conductance mapping
in Eq.~\eqref{eq:network-conductance-weight}. Thus, at
\(w_{\mathrm{norm}}=0.5\), the corresponding forward weights are \(0.5\)
and \(W_{\mathrm{fc1}}\approx0.023\), respectively.

All conditions used the same datasets, network sizes, batch sizes, spiking
settings, WTA dynamics, and 100-epoch schedule. Except for hidden-layer and
batch size, Hybrid settings were fixed to the dataset-specific values in
Table~\ref{tab:implementation-parameters}. Hybrid and Ideal STDP additionally
shared the local traces, STDP coefficients, batchwise update schedule, and
epoch-wise column normalization. Ideal STDP was evaluated only for the
1024-hidden-neuron, batch-128 configuration.

For the two backpropagation baselines, the fc2 learning rate, \(T\),
\(k\), and device parameters followed
Table~\ref{tab:implementation-parameters}. The remaining hyperparameters
were selected by final-epoch accuracy on a stratified validation hold-out
containing 15\% of the training data, using the 1024-hidden-neuron,
batch-128 configuration and a single seed. For Ideal BP, the fc1 Adam
learning rate was selected from
\(\{10^{-4},\,3\times10^{-4},\,10^{-3},\,3\times10^{-3}\}\), giving
\(3\times10^{-4}\) for both datasets. For Memristor BP, the fc1 Adam
learning rate was fixed at \(10^{-3}\) and the programming scale was
selected, because below the programming-coordinate limit the two enter the
update only as a product. Seven scales from \(4.10\times10^{4}\) to
\(1.01\times10^{7}\) were evaluated for MNIST and eleven from
\(7.30\times10^{2}\) to \(1.01\times10^{7}\) for Fashion-MNIST, with column
normalization enabled; the selected scales were \(4.10\times10^{4}\) and
\(7.30\times10^{2}\), respectively, the smallest values evaluated for each
dataset, and were also used without normalization. The selected settings were retrained on the full training
sets for all hidden-layer sizes, batch sizes, and seeds.

\begin{table*}[t]
\caption{Classification test accuracy under four fc1 learning
conditions: device-informed STDP (Hybrid), end-to-end backpropagation with
signed, unconstrained weights (Ideal BP), and backpropagation-driven
programming of the threshold-memristor states without and with epoch-wise
column normalization (Memristor BP and Memristor BP + norm.). Values are
mean \(\pm\) standard deviation over five seeds. Hybrid values are the
nominal Hybrid population. The comparison with Ideal STDP and the paired
statistical comparisons at the 1024-hidden-neuron, batch-128 configuration
are reported in the text.}
\label{tab:learning-rule-comparison}
\centering
\renewcommand{\arraystretch}{1.18}
\begin{tabular}{lllcccc}
\toprule
Dataset & Hidden & Batch & Hybrid & Ideal BP &
Memristor BP & Memristor BP + norm. \\
\midrule
Fashion-MNIST & 512  & 64  & $83.88\pm0.29$ & $87.68\pm0.55$ & $83.41\pm0.92$ & $66.04\pm6.81$ \\
Fashion-MNIST & 512  & 128 & $84.41\pm0.14$ & $87.99\pm0.24$ & $83.85\pm1.42$ & $76.91\pm1.85$ \\
Fashion-MNIST & 1024 & 64  & $85.34\pm0.17$ & $88.32\pm0.19$ & $82.36\pm3.41$ & $58.78\pm6.48$ \\
Fashion-MNIST & 1024 & 128 & $85.92\pm0.23$ & $88.37\pm0.33$ & $84.38\pm1.16$ & $81.51\pm2.51$ \\
MNIST         & 512  & 64  & $89.94\pm0.49$ & $97.79\pm0.32$ & $94.37\pm0.31$ & $67.11\pm17.35$ \\
MNIST         & 512  & 128 & $93.07\pm0.21$ & $97.93\pm0.14$ & $94.70\pm0.29$ & $87.49\pm6.28$ \\
MNIST         & 1024 & 64  & $93.74\pm0.04$ & $97.99\pm0.18$ & $94.81\pm0.41$ & $87.53\pm5.28$ \\
MNIST         & 1024 & 128 & $95.15\pm0.17$ & $97.78\pm0.44$ & $94.93\pm0.56$ & $82.87\pm14.97$ \\
\bottomrule
\end{tabular}
\end{table*}

Hybrid and Ideal STDP were compared in the seed-matched normalization
study described below. With column normalization, Hybrid reached
\(95.02\pm0.17\%\) on MNIST and \(84.79\pm0.17\%\) on Fashion-MNIST,
compared with \(92.33\pm0.35\%\) and \(83.25\pm0.22\%\) for Ideal STDP.
The paired Hybrid--Ideal-STDP differences were \(2.69\) percentage points
on MNIST (95\% CI \([2.28,3.10]\)) and \(1.54\) percentage points on
Fashion-MNIST (\([1.26,1.82]\)), with \(p_{\mathrm{Holm}}=1.1\times10^{-4}\)
for both; without normalization, they were \(3.56\) and \(1.67\)
percentage points.

Ideal BP reached \(97.78\)--\(97.99\%\) on MNIST and
\(87.68\)--\(88.37\%\) on Fashion-MNIST, exceeding Hybrid by
\(2.63\)--\(7.85\) and \(2.45\)--\(3.80\) percentage points, respectively
(Table~\ref{tab:learning-rule-comparison}). Memristor BP without
normalization reached \(94.37\)--\(94.93\%\) on MNIST and
\(82.36\)--\(84.38\%\) on Fashion-MNIST. Relative to the normalized Hybrid
values in Table~\ref{tab:learning-rule-comparison}, it was higher on MNIST in
three configurations and \(0.22\) percentage points lower at
1024 hidden neurons and batch 128; on Fashion-MNIST, Hybrid was higher in
all four configurations, by \(0.47\)--\(2.98\) percentage points.

Paired comparisons at the 1024-hidden-neuron, batch-128 configuration used
the five-seed Hybrid runs of the normalization study described below,
which share seeds and data subsets with the baselines; Memristor BP was
compared with Hybrid without normalization and Ideal BP with normalized
Hybrid. Two-sided paired \(t\)-tests were applied to the seed-level
final test accuracies, with Holm--Bonferroni adjustment within each
baseline. With normalization disabled, Hybrid and Memristor BP reached
\(94.31\pm0.25\%\) and \(94.93\pm0.56\%\) on MNIST (difference
\(-0.62\) percentage points, 95\% CI \([-1.53,0.30]\),
\(p_{\mathrm{Holm}}=0.27\)) and \(83.91\pm0.13\%\) and
\(84.38\pm1.16\%\) on Fashion-MNIST (\(-0.46\), \([-1.81,0.89]\),
\(p_{\mathrm{Holm}}=0.40\)). Relative to normalized Hybrid, Ideal BP was
higher by \(2.77\) percentage points on MNIST (95\% CI \([2.11,3.43]\),
\(p_{\mathrm{Holm}}=3.1\times10^{-4}\)) and \(3.58\) percentage points on
Fashion-MNIST (\([3.16,3.99]\), \(p_{\mathrm{Holm}}=3.7\times10^{-5}\)).

With epoch-wise column normalization, Memristor BP varied strongly across
seeds (Table~\ref{tab:learning-rule-comparison}). In the MNIST
512-hidden-neuron, batch-64 configuration, test accuracy ranged from
\(52.94\%\) to \(94.23\%\), while final training accuracy remained between
\(95.0\%\) and \(96.5\%\). Column normalization rescales the device states
after the last readout update of each epoch, so the readout is not
re-adapted to the final normalized fc1 before testing; the large
train--test gaps are consistent with this mismatch.

During validation, larger Memristor BP programming scales reduced
accuracy, particularly on Fashion-MNIST, where the highest-scale settings
drove device states to the conductance bounds. Because the
conductance-to-weight mapping is strongly nonlinear near
\(w_{\mathrm{norm}}=1\), a single saturated synapse can reach the unit
firing threshold, which degraded the hidden representation; the selected
Fashion-MNIST scale therefore produced comparatively small state
changes.

A fixed-random-fc1 control, with fc1 states initialized from
\(\mathcal{U}(0.3,0.7)\) and held fixed while fc2 was trained normally,
reached \(86.17\pm0.07\%\) on MNIST and \(79.15\pm0.07\%\) on
Fashion-MNIST over five seeds. A linear readout trained directly on the
rate-coded input spike counts, without fc1 but with the same data,
encoding, and readout training, reached \(92.65\pm0.15\%\) and
\(84.05\pm0.20\%\) over five seeds. The fixed random fc1 and direct
clipped Ideal STDP (\(92.33\pm0.35\%\) and \(83.25\pm0.22\%\)) fell
below this input-level readout, whereas the learned fc1 Hybrid
configuration exceeded it on both datasets, supporting a contribution
from the Hybrid fc1 representation.

A response-factor ablation was performed for the 1024-hidden-neuron,
batch-128 configuration. The constant-rate condition replaced the
resistance-, current-, and boundary-window-dependent
rate with polarity-specific reference constants evaluated at
\(w_{\mathrm{norm}}=0.5\) and \(V_{\mathrm{pulse}}=\pm0.6~\mathrm{V}\),
while preserving the reference SET/RESET directional-rate asymmetry. All
other training settings were held fixed, and each condition used five
matched seeds.

\begin{table}[t]
\caption{Ablation of the programming-rate response factor for the
1024-hidden-neuron, batch-128 configuration. The constant-rate condition
removes the resistance, current, and boundary-window dependence together.
Values are final test accuracy in percent, reported as
mean \(\pm\) standard deviation over five matched seeds.}
\label{tab:device-update-ablation}
\centering
\renewcommand{\arraystretch}{1.08}
\resizebox{\columnwidth}{!}{%
\begin{tabular}{lcc}
\hline
Condition & MNIST (\%) & Fashion-MNIST (\%) \\
\hline
Hybrid (full response factor)
& $95.15\pm0.17$ & $85.92\pm0.23$ \\
Constant rate
& $94.98\pm0.26$ & $85.63\pm0.09$ \\
\hline
\end{tabular}%
}
\end{table}

Replacing the resistance-, current-, and boundary-window-dependent response
factor with the constant-rate approximation reduced mean accuracy by
\(0.17\) percentage points on MNIST and \(0.29\) percentage points on
Fashion-MNIST. Because the nominal
\(\pm0.6~\mathrm{V}\) programming amplitudes exceed both adopted switching
thresholds, threshold gating is always satisfied at the nominal operating
point; threshold effects are therefore examined with the pulse-amplitude
sweep.

A three-seed pulse-amplitude sweep, trained on 51,000 images per dataset,
further separated the programming regimes
of the thresholded state equation. At each magnitude, positive and negative
STDP commands were applied as \(+V_{\mathrm{pulse}}\) and
\(-V_{\mathrm{pulse}}\) pulses, respectively, with \(S_{\mathrm{dur}}\) held at
its nominal value; the update magnitude therefore also varied with amplitude
through the branch currents. For
\(|V_{\mathrm{pulse}}|\leq0.15~\mathrm{V}\), both polarities remained
within the threshold interval, so fc1 received no programming updates and
changed only through the epoch-wise column normalization;
the resulting test accuracy was \(87.39\pm0.29\%\) on MNIST and
\(79.08\pm0.38\%\) on Fashion-MNIST. At \(0.20~\mathrm{V}\), only the
negative-polarity branch was active, yielding \(68.83\pm2.62\%\) and
\(76.04\pm0.12\%\), respectively. At \(0.30~\mathrm{V}\), the corresponding
accuracies were \(56.04\pm2.98\%\) and \(75.00\pm0.42\%\).
Single-polarity programming thus reduced classification accuracy below this
normalization-only baseline, whereas bidirectional programming at the nominal
amplitude improved on it.

The constant-rate ablation produced much smaller changes than the
\(2.69\)- and \(1.54\)-percentage-point paired Hybrid--Ideal-STDP differences.
Because it removes the resistance, current, and boundary-window dependence
together, it does not attribute the Hybrid--Ideal-STDP difference to any
single modeled term.

A separate \(2\times2\) study crossed the fc1 learning rule (Hybrid or
Ideal STDP) with column normalization on or off, using a separate five-seed
population with matched seeds across the four cells. On MNIST, normalization raised Hybrid from \(94.31\pm0.25\%\) to
\(95.02\pm0.17\%\) and Ideal STDP from \(90.74\pm0.39\%\) to
\(92.33\pm0.35\%\); on Fashion-MNIST, it raised Hybrid from
\(83.91\pm0.13\%\) to \(84.79\pm0.17\%\) and Ideal STDP from
\(82.25\pm0.30\%\) to \(83.25\pm0.22\%\). Normalization improved both rules,
and Hybrid remained higher with and without it.

\subsection{Sensitivity to Controlled STDP Update-Amplitude Asymmetry}
\label{subsec:stdp-amplitude-asymmetry}

LTP/LTD asymmetry is a recognized limitation of learning with resistive
synapses~\cite{burr2015experimental,kim2021snn}. Directional imbalance in the
local STDP command was introduced by scaling one
coefficient relative to the balanced value \(A_0\) by an asymmetry magnitude
\(\rho\geq1\):
\begin{equation}
(A_+,A_-)=
\begin{cases}
(\rho A_0,A_0), & \text{\(A_+\)-dominant},\\
(A_0,\rho A_0), & \text{\(A_-\)-dominant},
\end{cases}
\label{eq:rho}
\end{equation}
so that \(\rho=1\) recovers the balanced rule. Because one coefficient is scaled, \(\rho>1\) also increases the total update amplitude \(A_{+}+A_{-}\).
Both directions were evaluated
at matched \(\rho\) for the 1024-hidden-neuron, batch-128 configuration, with
the device-state equation, architecture, WTA setting, and fc2 readout
unchanged, over three seeds.

Device SET/RESET asymmetry and state-dependent update asymmetry are
intrinsic to the compact model and are characterized in
Section~\ref{subsec:device-characterization} and the ablation of
Section~\ref{subsec:learning-rule-comparison}; this experiment varies only
the algorithmic \(A_+/A_-\) asymmetry \(\rho\).

\begin{figure}[!t]
\centering
\includegraphics[width=\columnwidth]{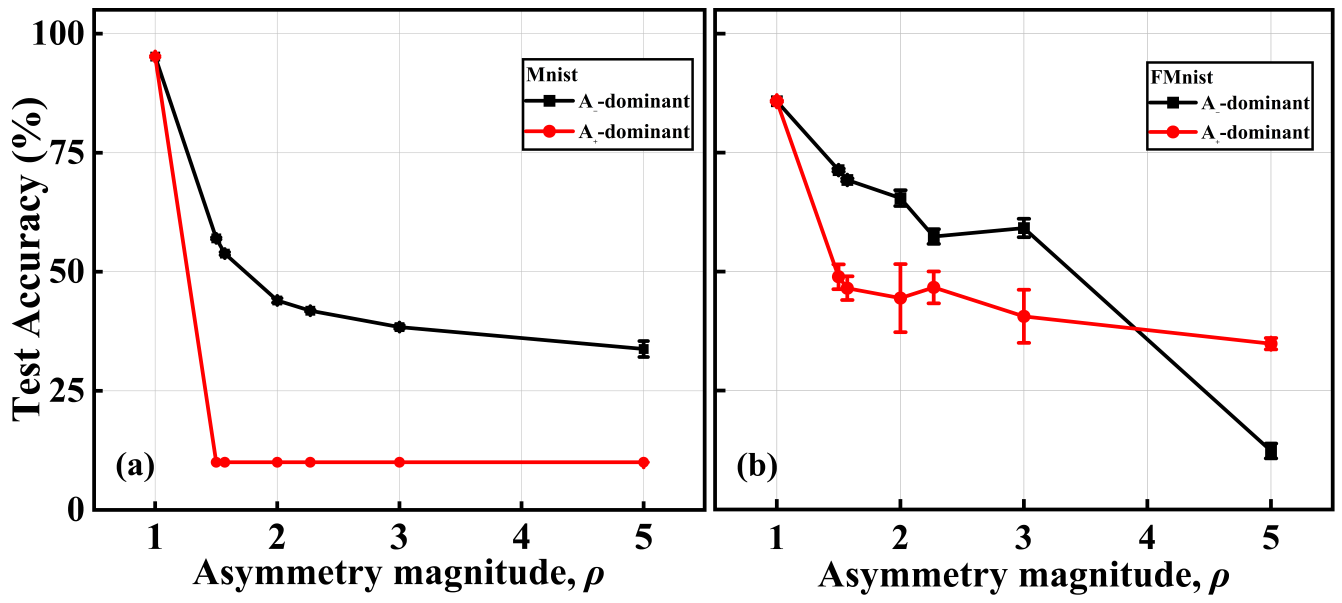}
\caption{Test accuracy versus STDP coefficient-asymmetry magnitude \(\rho\)
for the \(A_+\)- and \(A_-\)-dominant conditions of Eq.~\eqref{eq:rho}:
(a) MNIST and (b) Fashion-MNIST. Error bars denote one standard deviation
over three seeds.}
\label{fig:asymmetry-direction}
\end{figure}

Balanced coefficients gave the highest mean accuracy on both datasets
(Fig.~\ref{fig:asymmetry-direction}). The response to imbalance depended
strongly on direction. At \(\rho=1.5\), the \(A_+\)-dominant condition
fell to \(10.00\pm0.00\%\) on MNIST and \(48.91\pm3.20\%\) on
Fashion-MNIST, whereas the \(A_-\)-dominant condition retained
\(57.02\pm0.27\%\) and \(71.32\pm0.40\%\), respectively. On MNIST, the
\(A_+\)-dominant condition remained at chance level for all
\(\rho\geq1.5\), whereas the \(A_-\)-dominant condition decreased
gradually to \(33.79\pm2.05\%\) at \(\rho=5\). On Fashion-MNIST, the
ordering reversed at \(\rho=5\): the \(A_-\)-dominant condition fell to
\(12.31\pm1.92\%\), below the \(A_+\)-dominant \(34.84\pm1.47\%\).

A constant-sum control held \(A_{+}+A_{-}=2A_0\) fixed while varying
\(\rho\), separating the direction of imbalance from the total update
amplitude (three seeds, with a matched balanced reference of
\(95.06\pm0.09\%\) and \(84.80\pm0.19\%\)). The MNIST \(A_+\)-dominant
collapse persisted, with \(10.00\pm0.00\%\) at both \(\rho=1.5\) and
\(\rho=5\), whereas the \(A_-\)-dominant condition retained
\(58.78\pm0.27\%\) and \(41.08\pm0.17\%\); increasing the total update
amplitude was therefore not required for the collapse. On Fashion-MNIST,
the reversal at \(\rho=5\) did not persist when the total update amplitude
was held constant: the \(A_-\)-dominant condition reached
\(60.43\pm1.34\%\), above the \(A_+\)-dominant \(38.42\pm17.89\%\).

\subsubsection{Representation-Level Analysis of \texorpdfstring{\(A_{+}\)}{A+}-Dominant Collapse}
\label{subsubsec:asymmetry-mechanism}

The \(A_{+}\)-dominant MNIST performance collapse was characterized through
the final distribution of the normalized forward weight \(W_{\mathrm{fc1}}\)
and the resulting hidden representations.

\begin{figure}[!t]
\centering
\includegraphics[width=\columnwidth]{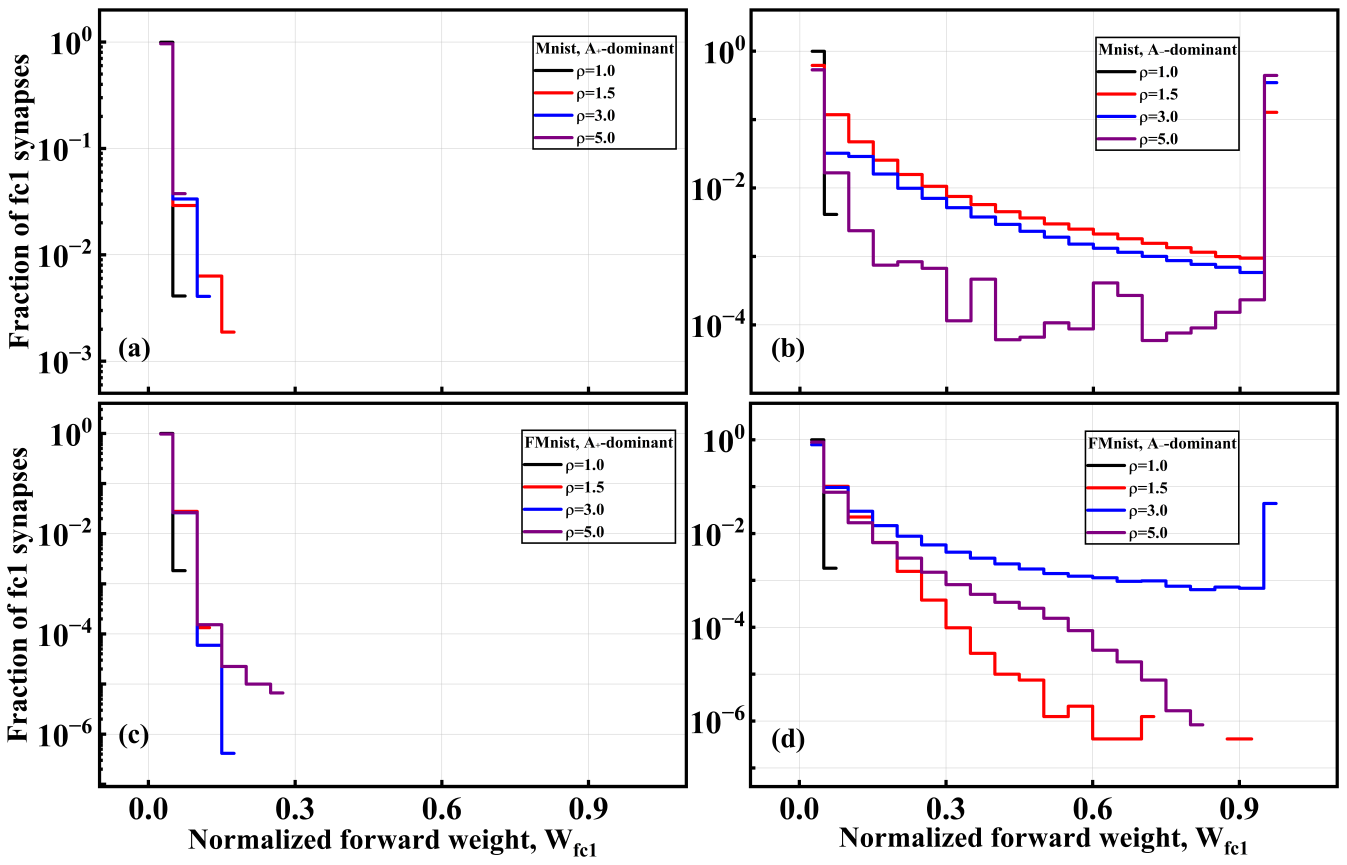}
\caption{Final fc1 effective-weight distributions under controlled STDP
coefficient-asymmetry magnitude \(\rho\). Each panel shows the mean fraction
of fc1 synapses in each bin of the conductance-derived forward weight
\(W_{\mathrm{fc1}}\) of Eq.~\eqref{eq:network-conductance-weight} over three seeds for
\(\rho=1.0\), \(1.5\), \(3.0\), and \(5.0\): (a) MNIST
\(A_+\)-dominant, (b) MNIST \(A_-\)-dominant, (c) Fashion-MNIST
\(A_+\)-dominant, and (d) Fashion-MNIST \(A_-\)-dominant. The vertical
axis is logarithmic; empty bins are not shown.}
\label{fig:asymmetry-weight-distributions}
\end{figure}

Figure~\ref{fig:asymmetry-weight-distributions}(a) shows that, under the
MNIST \(A_+\)-dominant condition at \(\rho=1.5\),
\(P(W_{\mathrm{fc1}}>0.95)\) was zero in every seed. The hidden
representation showed near-complete loss of input dependence. Different test images produced nearly identical hidden-layer spike-count
patterns, and in every seed all 8,920 test images were assigned to a single
output class, which differed across seeds. Because the subset is class balanced, this
constant prediction yields exactly the \(10.00\%\) accuracy reported above.
The same collapse occurred in every \(A_+\)-dominant MNIST run with
\(\rho\geq1.5\), with at least 99.9\% of test images assigned to a single class.

The mean fc1 firing rate remained at the WTA activity limit,
\(k/n_{\mathrm{hidden}}\approx0.0234\) spikes per hidden neuron per timestep,
for both the \(A_+\)-dominant \(\rho=1.5\) and balanced conditions. The
near-identical mean firing rates therefore reflect operation at the WTA
activity limit rather than preserved feature selectivity; the nearly
identical spike-count patterns across images instead show that the resulting
hidden representations became nearly input independent. In contrast, the
\(A_-\)-dominant \(\rho=1.5\) condition retained input-dependent features,
with clearly distinct hidden-layer patterns across images, a largest
predicted-class fraction of \(17.4\pm0.1\%\), and
\(57.02\pm0.27\%\) accuracy. Its \(W_{\mathrm{fc1}}\) distribution was
bimodal: \(12.76\pm0.03\%\) of weights exceeded 0.95, increasing to
\(44.28\pm0.21\%\) at \(\rho=5\), and the mean firing rate fell below the
WTA activity limit to 0.0169 spikes per hidden neuron per timestep. On
Fashion-MNIST, the \(A_-\)-dominant condition showed a smaller upper-bin
population at \(\rho=3\), where \(4.38\pm0.15\%\) of weights exceeded 0.95.

\subsection{Input Perturbation Response}
\label{subsec:input-perturbation-response}

\begin{figure}[!t]
\centering
\includegraphics[width=\columnwidth]{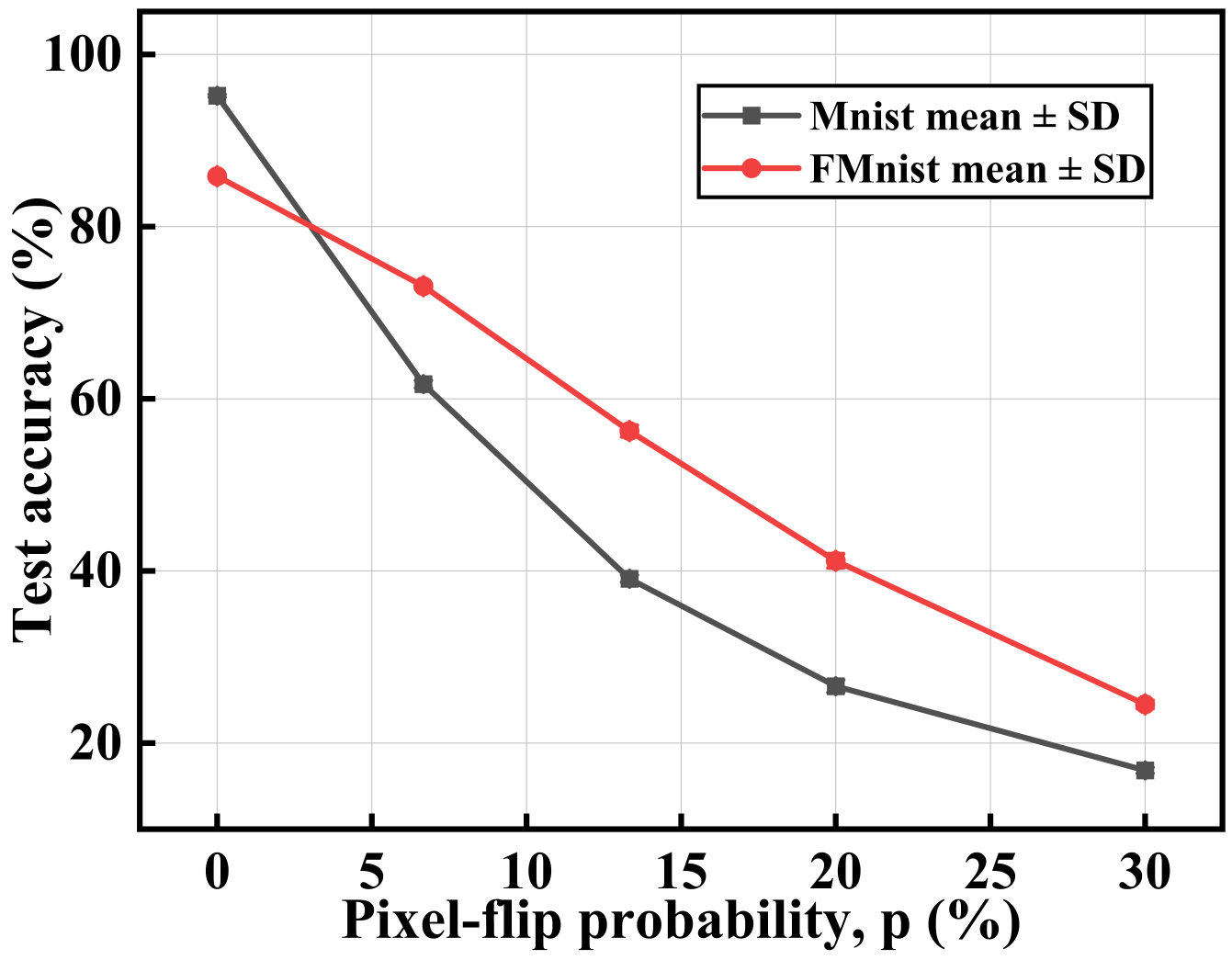}
\caption{Test accuracy under independent pixel-intensity inversion
(\(x_i\rightarrow1-x_i\)) at
\(p_{\mathrm{flip}}=\{0,6.67,13.33,20,30\}\%\). Error bars denote one standard deviation
over five independently trained models.}
\label{fig:input-noise-robustness}
\end{figure}

Input-perturbation sensitivity was evaluated by replacing each normalized
test-image pixel intensity \(x_i\) with \(1-x_i\) independently with
probability \(p_{\mathrm{flip}}\in\{0,6.67,13.33,20,30\}\%\), with the learned
fc1 states fixed during inference, using five independently trained
models per dataset. This isolates sensitivity to input corruption without
changes to the programmed synaptic states. Accuracy decreased monotonically with \(p_{\mathrm{flip}}\) on both datasets
(Fig.~\ref{fig:input-noise-robustness}). For MNIST, accuracy decreased from
\(95.21\pm0.17\%\) at \(p_{\mathrm{flip}}=0\) to \(61.71\pm0.43\%\),
\(39.10\pm0.41\%\), \(26.61\pm0.73\%\), and \(16.84\pm0.35\%\) at
\(p_{\mathrm{flip}}=6.67\%\), \(13.33\%\), \(20\%\), and \(30\%\), respectively. The
corresponding Fashion-MNIST accuracies were \(85.86\pm0.20\%\),
\(73.09\pm0.32\%\), \(56.26\pm0.68\%\), \(41.17\pm0.82\%\), and
\(24.51\pm0.49\%\).

The pixel-flip results are summarized in
Table~\ref{tab:robustness-summary}.

\subsection{Stuck-At Faults and Device-Parameter Variation}
\label{subsec:fault-variation-robustness}

Stuck-at faults are common defects in resistive crossbars, motivating
fault-aware training and mapping strategies~\cite{chen2017accelerator,
xia2017fault,liu2017rescuing}. Endpoint stuck-at faults were introduced before
training using an independently
sampled random mask for each run, with fault level denoting the fraction of
affected fc1 synapses; fault levels of 10\%, 20\%, and 30\% were
evaluated. Masked synapses were constrained to
\(w_{\mathrm{norm}}=0\) (stuck-at-\(G_{\min}\)) or
\(w_{\mathrm{norm}}=1\) (stuck-at-\(G_{\max}\)) throughout training and
evaluation; the assigned endpoint was reinstated after every local programming
update and after each epoch-wise column-normalization step. Faulted synapses
were included in the column sums used for normalization.

Device-to-device variation is a major nonideality in memristive
SNNs~\cite{querlioz2013immunity}. For device-parameter variation, each
fc1 synapse independently received
fixed Gaussian perturbations in \(R_{\mathrm{ON}}\), \(R_{\mathrm{OFF}}\),
\(V_{\mathrm{TH}}^{+}\), and \(V_{\mathrm{TH}}^{-}\), sampled once before training:
\begin{equation}
\theta'=\theta\left(1+\mathcal{N}(0,\sigma_{\mathrm{var}}^2)\right),
\label{eq:device-parameter-variation}
\end{equation}
where \(\sigma_{\mathrm{var}}\) is the relative standard deviation of the
Gaussian perturbation before clipping, applied independently to each
parameter.
Device-parameter variation was evaluated at
\(\sigma_{\mathrm{var}}\in\{0,5,10,20,25,30\}\%\), with
\(\sigma_{\mathrm{var}}=0\%\) providing the clean reference. Samples were clipped to
\(R_{\mathrm{ON}}\geq1\,\Omega\),
\(R_{\mathrm{OFF}}\geq10\,\Omega\),
\(V_{\mathrm{TH}}^{+}\geq10^{-3}\,\mathrm{V}\), and
\(V_{\mathrm{TH}}^{-}\leq-10^{-3}\,\mathrm{V}\), with
\(R_{\mathrm{OFF}}\geq1.5R_{\mathrm{ON}}\) enforced elementwise. The sampled
parameters remained fixed throughout each training and evaluation run.
Forward weights were computed with the nominal conductance bounds in
Eq.~\eqref{eq:network-conductance-weight} and clipped to \([0,1]\); at
\(\sigma_{\mathrm{var}}=30\%\), less than 1\% of weights were clipped.
Because clipping marks final states with \(R(w)>420~\Omega\), this provides
an upper bound on the fraction above the \(600~\Omega\) limit for
positive-branch programming at \(+0.6~\mathrm{V}\)
(Section~\ref{subsec:threshold-memristor-model}). At
\(\sigma_{\mathrm{var}}=30\%\), approximately 1.9\% of sampled
\(V_{\mathrm{TH}}^{+}\) values exceed \(0.6~\mathrm{V}\); these synapses
receive no potentiating updates.

This network-level variation experiment is distinct from the reduced HSPICE
crossbar mismatch study in Section~\ref{subsec:crossbar-mac-wire}, which uses
a fixed 5\% variation realization in \(R_{\mathrm{ON}}\) and
\(R_{\mathrm{OFF}}\) for static read-path analysis rather than the 5--30\%
four-parameter variation applied here during training and evaluation.

\begin{figure}[!t]
\centering
\includegraphics[width=\columnwidth]{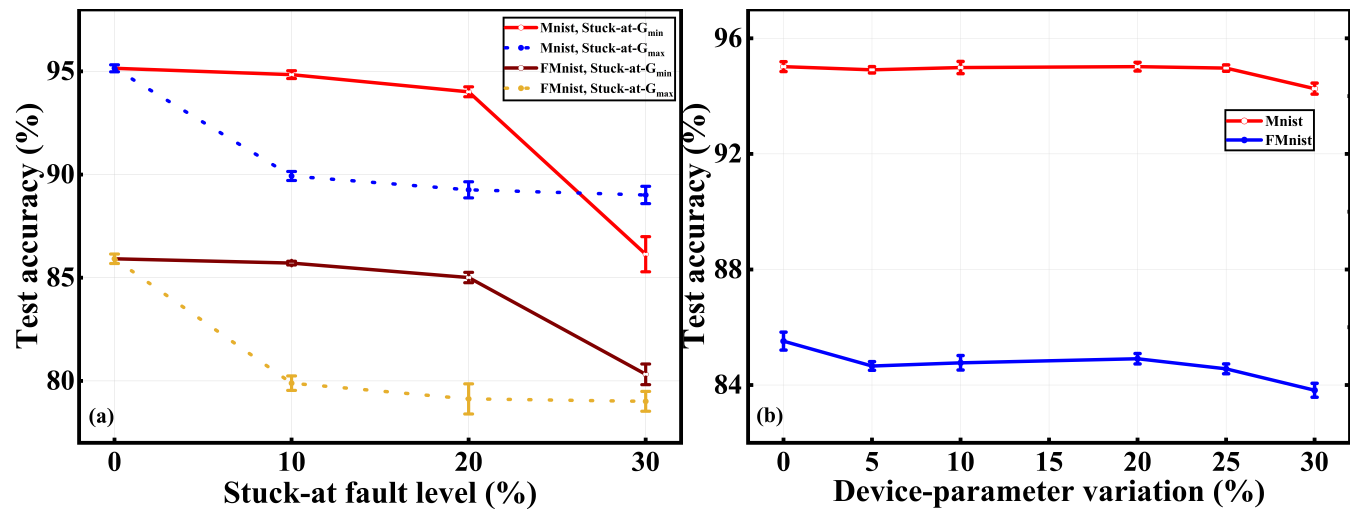}
\caption{Classification accuracy under (a) endpoint stuck-at faults present
throughout training and evaluation and (b) device-parameter variation in
\(R_{\mathrm{ON}}\), \(R_{\mathrm{OFF}}\), \(V_{\mathrm{TH}}^{+}\), and
\(V_{\mathrm{TH}}^{-}\). Variation levels denote the relative standard
deviation of the Gaussian perturbation before clipping.
Error bars denote one standard deviation over five seeds.}
\label{fig:fault-variation-robustness}
\end{figure}

Figure~\ref{fig:fault-variation-robustness} shows the effects of endpoint
faults present throughout training and of device-parameter variation.
Stuck-at-\(G_{\min}\) accuracy remained close to the matched clean reference
through 20\% faults, reaching \(94.01\pm0.24\%\) on MNIST and
\(85.01\pm0.25\%\) on Fashion-MNIST, before decreasing at 30\% to
\(86.14\pm0.85\%\) and \(80.32\pm0.50\%\), respectively.
Stuck-at-\(G_{\max}\) faults produced larger reductions at lower fault
levels, followed by comparatively small additional changes as the fault
fraction increased: accuracy was \(89.93\pm0.22\%\) on MNIST and
\(79.89\pm0.35\%\) on Fashion-MNIST at 10\%, and
\(89.01\pm0.42\%\) and \(79.01\pm0.48\%\), respectively, at 30\%.
These results are consistent with partial adaptation to endpoint faults
present during training. Under device-parameter variation, MNIST accuracy remained within
\(0.11\) percentage points of the clean reference through
\(\sigma_{\mathrm{var}}=25\%\), while Fashion-MNIST decreased by
\(0.61\)--\(0.96\) percentage points over the same range. At 30\%,
accuracy decreased to \(94.26\pm0.19\%\) on MNIST and
\(83.82\pm0.24\%\) on Fashion-MNIST.

\subsubsection{Post-Training Fault Injection}
\label{subsubsec:post-training-fault-injection}

To distinguish adaptation during learning from post-training resilience,
endpoint faults at 5\%, 10\%, and 30\% were injected into the final-epoch
clean-trained models of the preceding experiment without retraining, so that
neither fc1 nor the supervised readout could adapt. At the shared fault
levels of 10\% and 30\%, post-training injection produced substantially
larger accuracy reductions than faults present during training. At 5\%
stuck-at-\(G_{\max}\), MNIST accuracy decreased from
\(95.15\pm0.17\%\) to \(10.41\pm1.23\%\), and Fashion-MNIST accuracy
decreased from \(85.92\pm0.23\%\) to \(14.27\pm6.05\%\); accuracy
remained near chance at 10\% and 30\%. For stuck-at-\(G_{\min}\),
accuracy decreased to \(80.77\pm2.66\%\) on MNIST and
\(72.92\pm2.66\%\) on Fashion-MNIST at 5\%, to \(65.09\pm2.79\%\) and
\(61.25\pm6.59\%\) at 10\%, and to \(34.96\pm4.19\%\) and
\(38.03\pm3.98\%\) at 30\%. These smaller accuracy reductions when faults
were present during training are consistent with adaptation to defects
during learning, which post-training injection does not allow.

\begin{table*}[!t]
\centering
\caption{Summary of robustness experiments. Each clean reference is the
zero-perturbation result of the corresponding experiment; the two
endpoint-fault references are the nominal Hybrid population of
Table~\ref{tab:learning-rule-comparison}. Values are mean \(\pm\) standard
deviation in percent over five independently trained models (three for
write noise). Post-training
endpoint faults were injected into the final-epoch clean-trained models of the
training-time fault experiment without retraining. Changes are relative to the
clean reference within each experiment family.}
\label{tab:robustness-summary}
\renewcommand{\arraystretch}{1.08}
\setlength{\tabcolsep}{4.0pt}
\begin{tabular}{llrrrrl}
\toprule
\textbf{Experiment} &
\textbf{Condition} &
\multicolumn{2}{c}{\textbf{MNIST}} &
\multicolumn{2}{c}{\textbf{Fashion-MNIST}} &
\textbf{Timing} \\
\cmidrule(lr){3-4}
\cmidrule(lr){5-6}
&
&
\textbf{Accuracy (\%)} &
\textbf{Change (pp)} &
\textbf{Accuracy (\%)} &
\textbf{Change (pp)} &
\\
\midrule

Pixel-flip input noise &
$p_{\mathrm{flip}}=0$ &
$95.21\pm0.17$ &
--- &
$85.86\pm0.20$ &
--- &
After training \\

Pixel-flip input noise &
$p_{\mathrm{flip}}=6.67\%$ &
$61.71\pm0.43$ &
$-33.50$ &
$73.09\pm0.32$ &
$-12.77$ &
After training \\

Pixel-flip input noise &
$p_{\mathrm{flip}}=13.33\%$ &
$39.10\pm0.41$ &
$-56.11$ &
$56.26\pm0.68$ &
$-29.60$ &
After training \\

Pixel-flip input noise &
$p_{\mathrm{flip}}=20\%$ &
$26.61\pm0.73$ &
$-68.60$ &
$41.17\pm0.82$ &
$-44.69$ &
After training \\

Pixel-flip input noise &
$p_{\mathrm{flip}}=30\%$ &
$16.84\pm0.35$ &
$-78.37$ &
$24.51\pm0.49$ &
$-61.35$ &
After training \\

\midrule

Endpoint fault during training &
Clean ($0\%$) &
$95.15\pm0.17$ &
--- &
$85.92\pm0.23$ &
--- &
During training \\

Stuck-at-$G_{\min}$ &
$10\%$ &
$94.84\pm0.19$ &
$-0.31$ &
$85.71\pm0.09$ &
$-0.21$ &
During training \\

Stuck-at-$G_{\min}$ &
$20\%$ &
$94.01\pm0.24$ &
$-1.14$ &
$85.01\pm0.25$ &
$-0.91$ &
During training \\

Stuck-at-$G_{\min}$ &
$30\%$ &
$86.14\pm0.85$ &
$-9.01$ &
$80.32\pm0.50$ &
$-5.60$ &
During training \\

Stuck-at-$G_{\max}$ &
$10\%$ &
$89.93\pm0.22$ &
$-5.22$ &
$79.89\pm0.35$ &
$-6.03$ &
During training \\

Stuck-at-$G_{\max}$ &
$20\%$ &
$89.26\pm0.39$ &
$-5.89$ &
$79.13\pm0.73$ &
$-6.79$ &
During training \\

Stuck-at-$G_{\max}$ &
$30\%$ &
$89.01\pm0.42$ &
$-6.14$ &
$79.01\pm0.48$ &
$-6.91$ &
During training \\

\midrule

Device-parameter variation &
Clean ($0\%$) &
$95.02\pm0.17$ &
--- &
$85.52\pm0.31$ &
--- &
During training \\

Device-parameter variation &
$5\%$ &
$94.91\pm0.11$ &
$-0.11$ &
$84.66\pm0.15$ &
$-0.86$ &
During training \\

Device-parameter variation &
$10\%$ &
$94.99\pm0.21$ &
$-0.03$ &
$84.77\pm0.25$ &
$-0.75$ &
During training \\

Device-parameter variation &
$20\%$ &
$95.02\pm0.15$ &
$0.00$ &
$84.91\pm0.18$ &
$-0.61$ &
During training \\

Device-parameter variation &
$25\%$ &
$94.97\pm0.11$ &
$-0.05$ &
$84.56\pm0.17$ &
$-0.96$ &
During training \\

Device-parameter variation &
$30\%$ &
$94.26\pm0.19$ &
$-0.76$ &
$83.82\pm0.24$ &
$-1.70$ &
During training \\

\midrule

Post-training endpoint fault &
Clean ($0\%$) &
$95.15\pm0.17$ &
--- &
$85.92\pm0.23$ &
--- &
After training \\

Post-training stuck-at-$G_{\min}$ &
$5\%$ &
$80.77\pm2.66$ &
$-14.38$ &
$72.92\pm2.66$ &
$-13.00$ &
After training \\

Post-training stuck-at-$G_{\min}$ &
$10\%$ &
$65.09\pm2.79$ &
$-30.06$ &
$61.25\pm6.59$ &
$-24.67$ &
After training \\

Post-training stuck-at-$G_{\min}$ &
$30\%$ &
$34.96\pm4.19$ &
$-60.19$ &
$38.03\pm3.98$ &
$-47.89$ &
After training \\

Post-training stuck-at-$G_{\max}$ &
$5\%$ &
$10.41\pm1.23$ &
$-84.74$ &
$14.27\pm6.05$ &
$-71.65$ &
After training \\

Post-training stuck-at-$G_{\max}$ &
$10\%$ &
$9.85\pm2.29$ &
$-85.30$ &
$12.84\pm5.15$ &
$-73.08$ &
After training \\

Post-training stuck-at-$G_{\max}$ &
$30\%$ &
$10.65\pm1.78$ &
$-84.50$ &
$11.02\pm3.27$ &
$-74.90$ &
After training \\

\midrule

Update-proportional write noise &
$\sigma_{\mathrm{write}}=0$ &
$95.12\pm0.22$ &
--- &
$86.04\pm0.20$ &
--- &
During training \\

Update-proportional write noise &
$\sigma_{\mathrm{write}}=0.10$ &
$95.13\pm0.19$ &
$+0.01$ &
$85.92\pm0.15$ &
$-0.12$ &
During training \\

Update-proportional write noise &
$\sigma_{\mathrm{write}}=0.30$ &
$95.19\pm0.26$ &
$+0.07$ &
$85.86\pm0.09$ &
$-0.18$ &
During training \\

\bottomrule
\end{tabular}
\end{table*}

\subsection{Response to Controlled Update-Proportional Write Noise}
\label{subsec:write-noise-robustness}

Stochastic programming variability~\cite{boybat2018neuromorphic}, distinct
from the systematic directional
bias studied in Section~\ref{subsec:stdp-amplitude-asymmetry}, was modeled by
perturbing each normalized-state increment \(\Delta w\) before state clipping:
\begin{equation}
\begin{aligned}
\Delta w_{\mathrm{noisy}}
&=\Delta w+\epsilon,\\
\epsilon
&\sim\mathcal{N}\!\left(
0,
\left[
\sigma_{\mathrm{write}}\lvert\Delta w\rvert+10^{-12}
\right]^2
\right),
\end{aligned}
\label{eq:write-noise-model}
\end{equation}
where the dimensionless coefficient \(\sigma_{\mathrm{write}}\) sets the
perturbation standard deviation relative to \(\lvert\Delta w\rvert\), and the
\(10^{-12}\) term provides a numerical floor. An independent noise sample was
drawn for each fc1 synapse at every batchwise state update before clipping
the resulting normalized state to \([0,1]\). The nominal Hybrid configuration
uses \(\sigma_{\mathrm{write}}=0.018\), while
\(\sigma_{\mathrm{write}}=0\) provides a noise-free control.

Across
\(\sigma_{\mathrm{write}}\in\{0,0.018,0.049,0.10,0.20,0.30\}\), mean
accuracy over three seeds changed only slightly, remaining within \(95.12\)--\(95.22\%\) on
MNIST and \(85.84\)--\(86.04\%\) on Fashion-MNIST. Relative to the
noise-free \(\sigma_{\mathrm{write}}=0\) condition,
\(\sigma_{\mathrm{write}}=0.30\) changed mean accuracy by \(+0.07\) and
\(-0.18\) percentage points on MNIST and Fashion-MNIST, respectively.
These changes were substantially smaller than those observed under persistent
directional \(A_+/A_-\) imbalance. Other stochastic mechanisms, including
state-dependent or asymmetric noise, cycle-to-cycle drift, spatially
correlated variation, and retention loss, were not modeled and may produce
different sensitivity. Results are summarized in
Table~\ref{tab:robustness-summary}.

\section{Discussion and Conclusion}
\label{sec:discussion-conclusion}

This work examined how a thresholded, state-dependent memristor update
transforms local STDP learning in a hybrid SNN. The network-level mapping is a
validation-selected, reduced-order device-informed abstraction rather than a
pulse-accurate physical programming model. Local fc1 STDP commands
determined programming polarity and magnitude through a programming coordinate,
while the compact-model state equation determined the resulting
conductance-derived synaptic update. A supervised backpropagation-trained
fc2 readout performed class assignment.

In the learning-rule comparison, the Hybrid configuration achieved
\(95.15\pm0.17\%\) accuracy on the class-balanced MNIST test subsets and
\(85.92\pm0.23\%\) on Fashion-MNIST. In a separate seed-matched comparison,
Hybrid exceeded direct clipped Ideal STDP by 2.69 and 1.54 percentage
points on MNIST and Fashion-MNIST, respectively
(\(p_{\mathrm{Holm}}=1.1\times10^{-4}\)). A direct-input linear readout reached \(92.65\pm0.15\%\) on MNIST and
\(84.05\pm0.20\%\) on Fashion-MNIST, below the corresponding Hybrid
accuracies, supporting a contribution from the learned fc1
representation. The pulse-amplitude sweep further showed that single-polarity
programming reduced accuracy below the normalization-only baseline, whereas
bidirectional programming at the nominal amplitude improved on it.

Epoch-wise column normalization is an algorithmic homeostatic mechanism,
not a physical memristor operation. In the normalization ablation, enabling
normalization improved both Hybrid and direct clipped Ideal STDP, while Hybrid
remained higher with normalization disabled: \(94.31\pm0.25\%\) versus
\(90.74\pm0.39\%\) on MNIST and \(83.91\pm0.13\%\) versus
\(82.25\pm0.30\%\) on Fashion-MNIST. Implementation of this global
normalization would require an additional network- or circuit-level mechanism
outside the present hardware model. In the backpropagation comparison,
Ideal BP with signed, unconstrained weights exceeded Hybrid in every
configuration. Memristor BP, which programs the same device array, produced
accuracy similar to Hybrid at the 1024-hidden-neuron, batch-128
configuration without column normalization, while its normalized variant
showed larger seed-to-seed variation. Hybrid updates fc1 from local
pre- and post-synaptic traces, whereas Memristor BP derives its programming
commands from surrogate-gradient backpropagation.

Controlled STDP update-amplitude asymmetry produced severe,
direction-dependent accuracy degradation, and the MNIST \(A_+\)-dominant
collapse persisted with the total update amplitude held constant. At
\(\rho=1.5\),
the \(A_+\)-dominant condition decreased to \(10.00\pm0.00\%\) on MNIST and
\(48.91\pm3.20\%\) on Fashion-MNIST, whereas the corresponding
\(A_-\)-dominant condition retained \(57.02\pm0.27\%\) and
\(71.32\pm0.40\%\), respectively. For MNIST, representation-level analysis
showed that the \(A_+\)-dominant collapse was associated with nearly
input-independent hidden representations rather than reduced mean hidden-layer
firing or saturation of \(W_{\mathrm{fc1}}\) near its upper bound. Because the adopted
compact model exhibits state- and range-dependent SET/RESET behavior, the
coefficient sweep represents a controlled algorithmic \(A_+/A_-\) asymmetry
rather than a universal device-extracted SET/RESET ratio. Under the adopted
parameters, the model potentiation-to-depression rate ratio is approximately
7.3, 2.0, and 0.28 at \(w=0.2\), 0.5, and 0.8, respectively, and equals unity
near \(w\approx0.63\). The device mapping thus applies a polarity-specific,
state-dependent scaling to the STDP command, analogous to weight-dependent
(soft-bound) STDP~\cite{vanrossum2000stable,gutig2003learning}.

The endpoint-fault experiments distinguished adaptation during learning from
post-training resilience. When faults were present throughout training,
stuck-at-\(G_{\max}\) produced larger reductions at 10\% and 20\% fault
levels. At 30\%, the relative effect depended on dataset: stuck-at-\(G_{\min}\)
produced the larger reduction on MNIST (\(-9.01\) versus \(-6.14\) percentage
points), whereas stuck-at-\(G_{\max}\) produced the larger reduction on
Fashion-MNIST (\(-6.91\) versus \(-5.60\) percentage points). Post-training
injection produced substantially larger accuracy reductions, consistent with
partial adaptation to faults present during training. Stuck-at-\(G_{\max}\)
faults set affected synapses to the maximum conductance-derived forward
weight, \(W_{\mathrm{fc1}}=1\). At 5\%
stuck-at-\(G_{\max}\), accuracy decreased from \(95.15\pm0.17\%\) to
\(10.41\pm1.23\%\) on MNIST and from \(85.92\pm0.23\%\) to
\(14.27\pm6.05\%\) on Fashion-MNIST. Pixel-flip corruption also produced
monotonic accuracy degradation. In contrast, accuracy changed only slightly
across the tested update-proportional write-noise range
\(\sigma_{\mathrm{write}}\in\{0,0.018,0.049,0.10,0.20,0.30\}\), with
\(\sigma_{\mathrm{write}}=0.018\) corresponding to the nominal Hybrid
configuration.

The HSPICE analyses characterize the adopted compact model and reduced
differential-crossbar static read operation; they do not constitute a physical
implementation or energy evaluation of the complete \(784\times1024\)
network. At 0.6~V, the adopted resistance range
(\(R_{\mathrm{ON}}=10~\Omega\), \(R_{\mathrm{OFF}}=420~\Omega\)) corresponds
to device currents of 1.4--60~mA. Likewise, the network implementation uses
batchwise reduced-order
STDP-to-device updates rather than event-driven pulse-by-pulse programming.
Because the device dynamics are state dependent, sequential event-level
programming can in principle produce trajectories different from an aggregated
update evaluated at the pre-update state. Evaluation at batch sizes of 64 and
128 therefore does not establish equivalence to single-sample or
spike-event-level device programming.

Overall, under the evaluated conditions, incorporating thresholded,
state-dependent programming into local SNN learning improved accuracy over
direct clipped STDP in seed-matched comparisons, and the resulting networks
showed sensitivity that differed strongly across the controlled
nonidealities considered here. Future work should extend the framework to
experimentally calibrated device distributions, sequential event-driven
programming, array-scale parasitic models, realistic write scheduling and
peripheral circuits, and energy--latency--accuracy evaluation of complete
hardware implementations.

\bibliographystyle{IEEEtran}
\bibliography{reference3}

@article{mead1990neuromorphic,
  author  = {Carver Mead},
  title   = {Neuromorphic Electronic Systems},
  journal = {Proceedings of the IEEE},
  volume  = {78},
  number  = {10},
  pages   = {1629--1636},
  year    = {1990},
  doi     = {10.1109/5.58356}
}

@article{maass1997networks,
  author  = {Wolfgang Maass},
  title   = {Networks of Spiking Neurons: The Third Generation of Neural Network Models},
  journal = {Neural Networks},
  volume  = {10},
  number  = {9},
  pages   = {1659--1671},
  year    = {1997},
  doi     = {10.1016/S0893-6080(97)00011-7}
}

@article{indiveri2015memory,
  author  = {Giacomo Indiveri and Shih-Chii Liu},
  title   = {Memory and Information Processing in Neuromorphic Systems},
  journal = {Proceedings of the IEEE},
  volume  = {103},
  number  = {8},
  pages   = {1379--1397},
  year    = {2015},
  doi     = {10.1109/JPROC.2015.2444094}
}

@article{furber2016large,
  author  = {Steve B. Furber},
  title   = {Large-Scale Neuromorphic Computing Systems},
  journal = {Journal of Neural Engineering},
  volume  = {13},
  number  = {5},
  pages   = {051001},
  year    = {2016},
  doi     = {10.1088/1741-2560/13/5/051001}
}

@article{strukov2008missing,
  author  = {Dmitri B. Strukov and Gregory S. Snider and Duncan R. Stewart and R. Stanley Williams},
  title   = {The Missing Memristor Found},
  journal = {Nature},
  volume  = {453},
  number  = {7191},
  pages   = {80--83},
  year    = {2008},
  doi     = {10.1038/nature06932}
}

@article{burr2017neuromorphic,
  author  = {Geoffrey W. Burr and Robert M. Shelby and Abu Sebastian and
             Sangbum Kim and Seyoung Kim and Severin Sidler and Kumar Virwani and
             Masatoshi Ishii and Pritish Narayanan and Alessandro Fumarola and
             Lucas L. Sanches and Irem Boybat and Manuel {Le Gallo} and
             Kibong Moon and Jiyoo Woo and Hyunsang Hwang and Yusuf Leblebici},
  title   = {Neuromorphic Computing Using Non-Volatile Memory},
  journal = {Advances in Physics: X},
  volume  = {2},
  number  = {1},
  pages   = {89--124},
  year    = {2017},
  doi     = {10.1080/23746149.2016.1259585}
}

@article{ielmini2018inmemory,
  author  = {Daniele Ielmini and H.-S. Philip Wong},
  title   = {In-Memory Computing with Resistive Switching Devices},
  journal = {Nature Electronics},
  volume  = {1},
  pages   = {333--343},
  year    = {2018},
  doi     = {10.1038/s41928-018-0092-2}
}

@article{sebastian2020memory,
  author  = {Abu Sebastian and Manuel {Le Gallo} and Riduan Khaddam-Aljameh and Evangelos Eleftheriou},
  title   = {Memory Devices and Applications for In-Memory Computing},
  journal = {Nature Nanotechnology},
  volume  = {15},
  pages   = {529--544},
  year    = {2020},
  doi     = {10.1038/s41565-020-0655-z}
}

@article{serrano2013stdp,
  author  = {Teresa Serrano-Gotarredona and Timoth{\'e}e Masquelier and Themistoklis Prodromakis and Giacomo Indiveri and Bernabe Linares-Barranco},
  title   = {{STDP} and {STDP} Variations with Memristors for Spiking Neuromorphic Learning Systems},
  journal = {Frontiers in Neuroscience},
  volume  = {7},
  pages   = {2},
  year    = {2013},
  doi     = {10.3389/fnins.2013.00002}
}

@article{querlioz2013immunity,
  author  = {Damien Querlioz and Olivier Bichler and Philippe Dollfus and Christian Gamrat},
  title   = {Immunity to Device Variations in a Spiking Neural Network with Memristive Nanodevices},
  journal = {IEEE Transactions on Nanotechnology},
  volume  = {12},
  number  = {3},
  pages   = {288--295},
  year    = {2013},
  doi     = {10.1109/TNANO.2013.2250995}
}

@article{kim2021snn,
  author  = {Taeyoon Kim and Suman Hu and Jaewook Kim and Joon Young Kwak and
             Jongkil Park and Suyoun Lee and Inho Kim and Jong-Keuk Park and
             YeonJoo Jeong},
  title   = {Spiking Neural Network ({SNN}) with Memristor Synapses Having Non-Linear Weight Update},
  journal = {Frontiers in Computational Neuroscience},
  volume  = {15},
  pages   = {646125},
  year    = {2021},
  doi     = {10.3389/fncom.2021.646125}
}

@inproceedings{agarwal2016resistive,
  author    = {Sapan Agarwal and others},
  title     = {Resistive Memory Device Requirements for a Neural Algorithm Accelerator},
  booktitle = {Proc. International Joint Conference on Neural Networks (IJCNN)},
  pages     = {929--938},
  year      = {2016},
  doi       = {10.1109/IJCNN.2016.7727298}
}

@article{sayyaparaju2020deviceaware,
  author  = {Sagarvarma Sayyaparaju and Md. Musabbir Adnan and Sherif Amer and Garrett S. Rose},
  title   = {Device-Aware Circuit Design for Robust Memristive Neuromorphic Systems with {STDP}-Based Learning},
  journal = {ACM Journal on Emerging Technologies in Computing Systems},
  volume  = {16},
  number  = {3},
  pages   = {28:1--28:25},
  year    = {2020},
  doi     = {10.1145/3380969}
}

@article{sung2018review,
  author  = {Changhwan Sung and Hyunsang Hwang and In Kyeong Yoo},
  title   = {Perspective: A Review on Memristive Hardware for Neuromorphic Computation},
  journal = {Journal of Applied Physics},
  volume  = {124},
  number  = {15},
  pages   = {151903},
  year    = {2018},
  doi     = {10.1063/1.5044638}
}

@article{wang2025realtime,
  author  = {Zixu Wang and Wenhao Song and Tong Wang and Zihan Wang and
             Yichun Xu and Mingyi Rao and Fuxi Cai and Wenbo Yin and
             Mike Shuo-Wei Chen and Ning Ge and Maxwell Collins and
             Kangjun Bai and Sabyasachi Ganguli and Michael R. Page and
             Qing Wu and Linda Katehi and Qiangfei Xia and Miao Hu and
             J. Joshua Yang},
  title   = {Real-Time Signal Processing Enabled by Fused Networks on a
             Memristor-Based System on a Chip},
  journal = {Science Advances},
  volume  = {11},
  number  = {30},
  pages   = {eadv3436},
  year    = {2025},
  doi     = {10.1126/sciadv.adv3436}
}

@misc{prajapati2025efficient,
  author       = {Santlal Prajapati and Susmita Sur-Kolay and Soumyadeep Dutta},
  title        = {Efficient Memristive Spiking Neural Networks Architecture with Supervised In-Situ {STDP} Method},
  howpublished = {arXiv preprint arXiv:2507.20998},
  year         = {2025},
  url          = {https://arxiv.org/abs/2507.20998}
}

@article{zhang2017memristive,
  author  = {Yang Zhang and Xiaoping Wang and Yi Li and Eby G. Friedman},
  title   = {Memristive Model for Synaptic Circuits},
  journal = {IEEE Transactions on Circuits and Systems II: Express Briefs},
  volume  = {64},
  number  = {7},
  pages   = {767--771},
  year    = {2017},
  doi     = {10.1109/TCSII.2016.2605069}
}

@article{jo2010nanoscale,
  author  = {Sung Hyun Jo and Ting Chang and Idongesit Ebong and Bhavitavya B. Bhadviya and Pinaki Mazumder and Wei Lu},
  title   = {Nanoscale Memristor Device as Synapse in Neuromorphic Systems},
  journal = {Nano Letters},
  volume  = {10},
  number  = {4},
  pages   = {1297--1301},
  year    = {2010},
  doi     = {10.1021/nl904092h}
}

@article{prezioso2015training,
  author  = {Mirko Prezioso and Farnood Merrikh-Bayat and Brian D. Hoskins and Gina C. Adam and Konstantin K. Likharev and Dmitri B. Strukov},
  title   = {Training and Operation of an Integrated Neuromorphic Network Based on Metal-Oxide Memristors},
  journal = {Nature},
  volume  = {521},
  number  = {7550},
  pages   = {61--64},
  year    = {2015},
  doi     = {10.1038/nature14441}
}

@article{lecun1998gradient,
  author  = {Yann LeCun and L{\'e}on Bottou and Yoshua Bengio and Patrick Haffner},
  title   = {Gradient-Based Learning Applied to Document Recognition},
  journal = {Proceedings of the IEEE},
  volume  = {86},
  number  = {11},
  pages   = {2278--2324},
  year    = {1998},
  doi     = {10.1109/5.726791}
}

@misc{xiao2017fashion,
  author       = {Han Xiao and Kashif Rasul and Roland Vollgraf},
  title        = {{Fashion-MNIST}: A Novel Image Dataset for Benchmarking Machine Learning Algorithms},
  howpublished = {arXiv preprint arXiv:1708.07747},
  year         = {2017}
}

@article{neftci2019surrogate,
  author  = {Emre O. Neftci and Hesham Mostafa and Friedemann Zenke},
  title   = {Surrogate Gradient Learning in Spiking Neural Networks: Bringing the Power of Gradient-Based Optimization to Spiking Neural Networks},
  journal = {IEEE Signal Processing Magazine},
  volume  = {36},
  number  = {6},
  pages   = {51--63},
  year    = {2019},
  doi     = {10.1109/MSP.2019.2931595}
}

@article{burr2015experimental,
  author  = {Geoffrey W. Burr and Robert M. Shelby and Severin Sidler and
             Carmelo di Nolfo and Junwoo Jang and Irem Boybat and
             Rohit S. Shenoy and Pritish Narayanan and Kumar Virwani and
             Emanuele U. Giacometti and Bulent N. Kurdi and Hyunsang Hwang},
  title   = {Experimental Demonstration and Tolerancing of a Large-Scale
             Neural Network (165\,000 Synapses) Using Phase-Change Memory
             as the Synaptic Weight Element},
  journal = {IEEE Transactions on Electron Devices},
  volume  = {62},
  number  = {11},
  pages   = {3498--3507},
  year    = {2015},
  doi     = {10.1109/TED.2015.2439635}
}

@article{yu2018neuroinspired,
  author  = {Shimeng Yu},
  title   = {Neuro-Inspired Computing With Emerging Nonvolatile Memorys},
  journal = {Proceedings of the IEEE},
  volume  = {106},
  number  = {2},
  pages   = {260--285},
  year    = {2018},
  doi     = {10.1109/JPROC.2018.2790840}
}

@article{pedretti2017memristive,
  author  = {Giacomo Pedretti and Valerio Milo and Stefano Ambrogio and Roberto Carboni
             and Stefano Bianchi and Alessandro Calderoni and Nirmal Ramaswamy
             and Alessandro S. Spinelli and Daniele Ielmini},
  title   = {Memristive Neural Network for On-Line Learning and Tracking
             with Brain-Inspired Spike Timing Dependent Plasticity},
  journal = {Scientific Reports},
  volume  = {7},
  pages   = {5288},
  year    = {2017},
  doi     = {10.1038/s41598-017-05480-0}
}

@article{prezioso2018spike,
  author  = {Mirko Prezioso and M. Reza Mahmoodi and Farnood Merrikh Bayat
             and Hussein Nili and Hyungjin Kim and Adrien Vincent and Dmitri B. Strukov},
  title   = {Spike-Timing-Dependent Plasticity Learning of Coincidence
             Detection with Passively Integrated Memristive Circuits},
  journal = {Nature Communications},
  volume  = {9},
  pages   = {5311},
  year    = {2018},
  doi     = {10.1038/s41467-018-07757-y}
}

@article{wang2018fully,
  author  = {Zhongrui Wang and Saumil Joshi and Sergey Savel'ev
             and Wenhao Song and Rivu Midya and Yunning Li and Mingyi Rao
             and Peng Yan and Shiva Asapu and Ye Zhuo and Hao Jiang
             and Peng Lin and Can Li and Jung Ho Yoon
             and Navnidhi K. Upadhyay and Jiaming Zhang and Miao Hu
             and John Paul Strachan and Mark Barnell and Qing Wu
             and Huaqiang Wu and R. Stanley Williams and Qiangfei Xia
             and J. Joshua Yang},
  title   = {Fully Memristive Neural Networks for Pattern Classification
             with Unsupervised Learning},
  journal = {Nature Electronics},
  volume  = {1},
  pages   = {137--145},
  year    = {2018},
  doi     = {10.1038/s41928-018-0023-2}
}

@article{joglekar2009elusive,
  author  = {Yogesh N. Joglekar and Stephen J. Wolf},
  title   = {The Elusive Memristor: Properties of Basic Electrical Circuits},
  journal = {European Journal of Physics},
  volume  = {30},
  number  = {4},
  pages   = {661--675},
  year    = {2009},
  doi     = {10.1088/0143-0807/30/4/001}
}

@article{liang2010crosspoint,
  author  = {Jiale Liang and H.-S. Philip Wong},
  title   = {Cross-Point Memory Array Without Cell Selectors---Device
             Characteristics and Data Storage Pattern Dependencies},
  journal = {IEEE Transactions on Electron Devices},
  volume  = {57},
  number  = {10},
  pages   = {2531--2538},
  year    = {2010},
  doi     = {10.1109/TED.2010.2062187}
}

@article{chen2013comprehensive,
  author  = {An Chen},
  title   = {A Comprehensive Crossbar Array Model with Solutions for
             Line Resistance and Nonlinear Device Characteristics},
  journal = {IEEE Transactions on Electron Devices},
  volume  = {60},
  number  = {4},
  pages   = {1318--1326},
  year    = {2013},
  doi     = {10.1109/TED.2013.2246791}
}

@article{diehl2015unsupervised,
  author  = {Peter U. Diehl and Matthew Cook},
  title   = {Unsupervised Learning of Digit Recognition Using
             Spike-Timing-Dependent Plasticity},
  journal = {Frontiers in Computational Neuroscience},
  volume  = {9},
  pages   = {99},
  year    = {2015},
  doi     = {10.3389/fncom.2015.00099}
}

@article{zenke2018superspike,
  author  = {Friedemann Zenke and Surya Ganguli},
  title   = {{SuperSpike}: Supervised Learning in Multilayer Spiking Neural Networks},
  journal = {Neural Computation},
  volume  = {30},
  number  = {6},
  pages   = {1514--1541},
  year    = {2018},
  doi     = {10.1162/neco_a_01086}
}

@inproceedings{chen2017accelerator,
  title={Accelerator-friendly neural-network training: Learning variations and defects in RRAM crossbar},
  author={Chen, Lerong and Li, Jiawen and Chen, Yiran and Deng, Qiuping and Shen, Jiyuan and Liang, Xiaoyao and Jiang, Li},
  booktitle={Design, Automation \& Test in Europe Conference \& Exhibition (DATE), 2017},
  pages={19--24},
  year={2017},
  organization={IEEE}
}

@inproceedings{xia2017fault,
  title={Fault-tolerant training with on-line fault detection for RRAM-based neural computing systems},
  author={Xia, Lixue and Liu, Mengyun and Ning, Xuefei and Chakrabarty, Krishnendu and Wang, Yu},
  booktitle={Proceedings of the 54th Annual Design Automation Conference 2017},
  pages={1--6},
  year={2017}
}

@inproceedings{liu2017rescuing,
  author    = {Chenchen Liu and Miao Hu and John Paul Strachan and Hai Helen Li},
  title     = {Rescuing Memristor-Based Neuromorphic Design with High Defects},
  booktitle = {Proc. 54th Annual Design Automation Conf. (DAC)},
  year      = {2017},
  doi       = {10.1145/3061639.3062310}
}

@article{boybat2018neuromorphic,
  author  = {Irem Boybat and Manuel {Le Gallo} and S. R. Nandakumar and
             Timoleon Moraitis and Thomas Parnell and Tomas Tuma and
             Bipin Rajendran and Yusuf Leblebici and Abu Sebastian and
             Evangelos Eleftheriou},
  title   = {Neuromorphic Computing with Multi-Memristive Synapses},
  journal = {Nature Communications},
  volume  = {9},
  pages   = {2514},
  year    = {2018},
  doi     = {10.1038/s41467-018-04933-y}
}

@article{zhang2026hardware,
  author  = {Ouwen Zhang and Dainan Zhang and Junjie Wang and Shuang Liu and
             Hao Jiang and Zhongrui Wang and Xiaojuan Qi},
  title   = {A Memristor-Based Spiking Neural Network Circuit with Hardware-Optimized Unsupervised {STDP}},
  journal = {Microelectronics Journal},
  volume  = {167},
  pages   = {106916},
  year    = {2026},
  doi     = {10.1016/j.mejo.2025.106916}
}

@inproceedings{vu2024nonidealities,
  author    = {Viet Cuong Vu and Anthony Kenyon and Dovydas Joksas and Adnan Mehonic and Daniel J. Mannion and Wing H. Ng},
  title     = {Spiking Neural Networks with Nonidealities from Memristive Silicon Oxide Devices},
  booktitle = {Proc. IEEE 24th International Conference on Nanotechnology (NANO)},
  pages     = {46--50},
  year      = {2024},
  doi       = {10.1109/NANO61778.2024.10628612}
}

@article{vanrossum2000stable,
  author  = {van Rossum, M. C. W. and Bi, G. Q. and Turrigiano, G. G.},
  title   = {Stable {Hebbian} learning from spike timing-dependent plasticity},
  journal = {J. Neurosci.},
  volume  = {20}, number = {23}, pages = {8812--8821}, year = {2000}
}

@article{gutig2003learning,
  author  = {G{\"u}tig, R. and Aharonov, R. and Rotter, S. and Sompolinsky, H.},
  title   = {Learning input correlations through nonlinear temporally asymmetric {Hebbian} plasticity},
  journal = {J. Neurosci.},
  volume  = {23}, number = {9}, pages = {3697--3714}, year = {2003}
}
\end{document}